\pdfoutput=1
\documentclass[11pt]{article}

\usepackage[preprint]{acl}

\usepackage{times}
\usepackage{latexsym}
\usepackage{tabularx}
\usepackage{booktabs}
\usepackage{inconsolata}
\usepackage{subcaption}
\usepackage{xcolor}
\usepackage{booktabs}
\usepackage{multirow}
\usepackage{caption}
\usepackage{adjustbox}
\usepackage{amsmath}
\usepackage{amsthm}
\usepackage{bm}
\usepackage{float}

\usepackage[T1]{fontenc}
\usepackage[utf8]{inputenc}

\usepackage{microtype}

\usepackage{inconsolata}

\usepackage{graphicx}

\title{\includegraphics[scale=0.03]{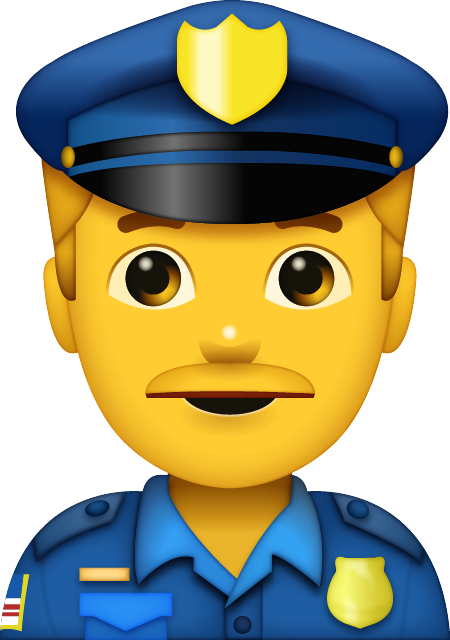} CAP: Data Contamination Detection via Consistency Amplification}

\author{
    Yi Zhao\textsuperscript{\rm 1}, 
    Jing Li\textsuperscript{\rm 1} ,
    Linyi Yang\textsuperscript{\rm 2, \thanks{Corresponding author}}\\
    \textsuperscript{1} Department of Computing, The Hong Kong Polytechnic University \\
    \textsuperscript{2} School of Engineering, Westlake University \\
    \texttt{23037086r@connect.polyu.hk}, \texttt{jing-amelia.li@polyu.edu.hk} \\
    \texttt{yanglinyi@westlake.edu.cn} \\
}

\begin{document}
\maketitle
\begin{abstract}
Large language models (LLMs) are widely used, but concerns about data contamination challenge the reliability of LLM evaluations.
Existing contamination detection methods are often task-specific or require extra prerequisites, limiting practicality.
We propose a novel framework, Consistency Amplification-based Data Contamination Detection (CAP), which introduces the Performance Consistency Ratio (PCR) to measure dataset leakage by leveraging LM consistency.
To the best of our knowledge, this is the first method to explicitly differentiate between fine-tuning and contamination, which is crucial for detecting contamination in domain-specific models.
Additionally, CAP is applicable to various benchmarks and works for both white-box and black-box models.
We validate CAP's effectiveness through experiments on seven LLMs and four domain-specific benchmarks.
Our findings also show that composite benchmarks from various dataset sources are particularly prone to unintentional contamination.
Codes will be publicly available soon.
\end{abstract}

\section{Introduction}
Large language models (LLMs) have significantly advanced natural language processing (NLP), leading to widespread use in applications such as translation \citep{DBLP:conf/acl/AgostinelliWRFC24}, summarization \citep{DBLP:conf/iclr/ChangLGI24} and question answering \citep{DBLP:conf/acl/SchimanskiNKAL24}. Notable models include GPT \citep{openai2024gpt4technicalreport} series and LLaMA \citep{touvron2023llama2openfoundation}. Beyond general-purpose applications, some LLMs have been fine-tuned for specific domains, producing specialized models with expertise in fields like law \citep{cui2024chatlawmultiagentcollaborativelegal}, biomedicine \citep{bolton2024biomedlm27bparameterlanguage}, finance \citep{yang2023investlmlargelanguagemodel}, and software development \citep{chen2021evaluatinglargelanguagemodels}.
\begin{figure}[ht]
    \centering
    \begin{subfigure}[t]{\columnwidth}
        \centering
        \includegraphics[width=\textwidth]{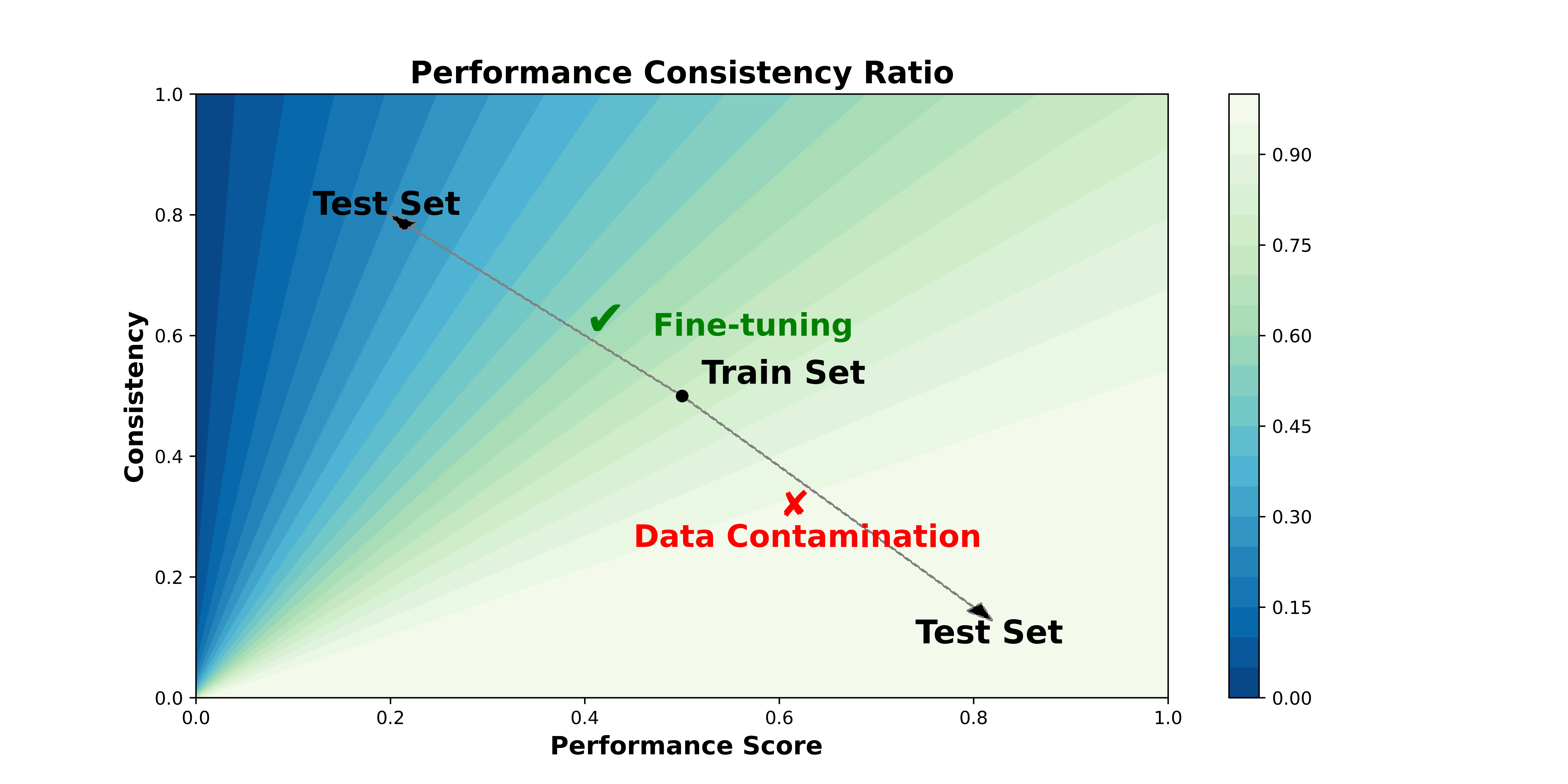}
        \caption{PCR Value Heatmap}
        \label{fig:heatmap}
    \end{subfigure}
    \begin{subfigure}[t]{\columnwidth}
        \centering
        \includegraphics[width=0.9\textwidth]{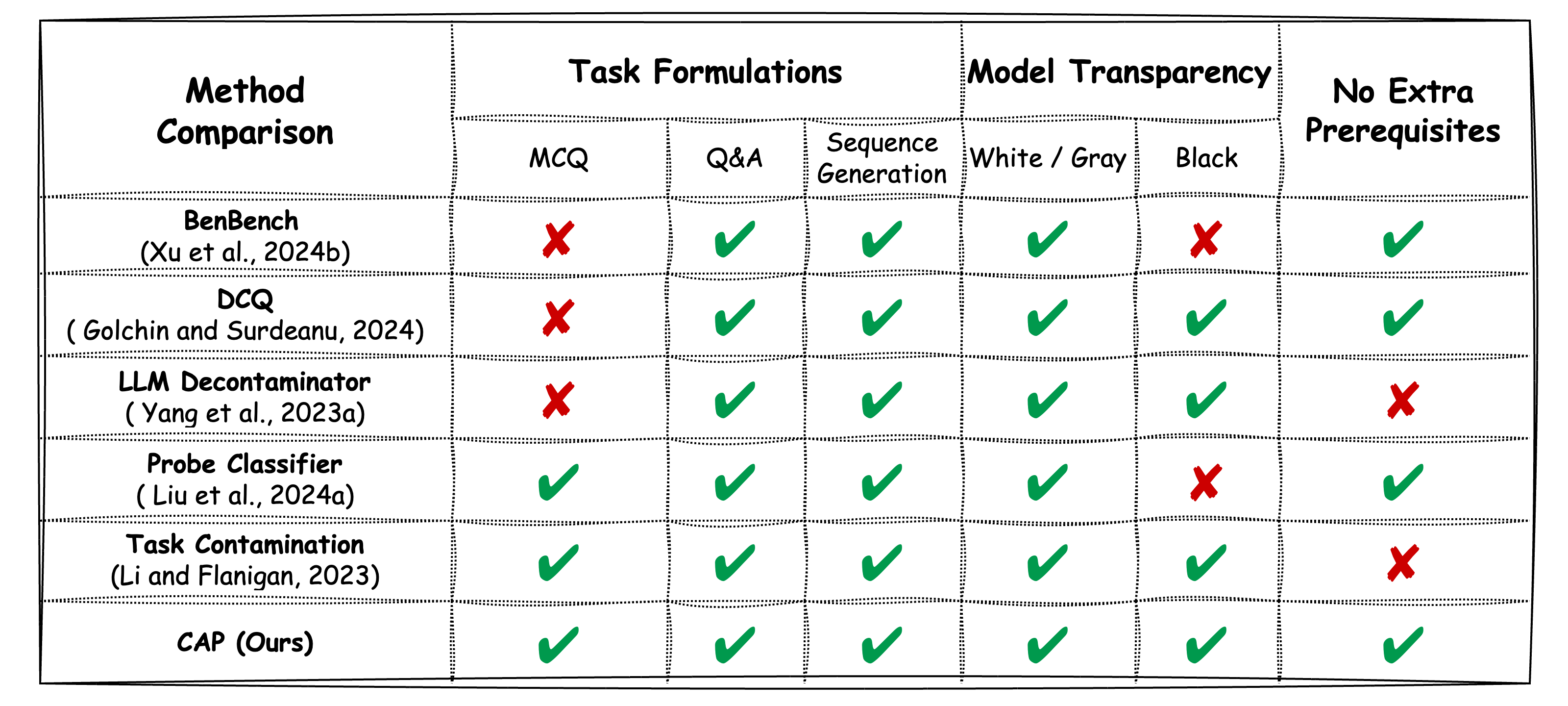}
        \caption{Method Comparison}
        \label{fig:comparison}
    \end{subfigure}
    \caption{(a) Heatmap of Performance Consistency Ratios: Change direction indicates scenario: \textit{Fine-Tuning} or \textit{Data Contamination}. (b) Method Comparison: CAP supports diverse tasks and is effective across models at varying transparency levels; it requires no prerequisites.}
    \label{fig:heatmap_and_comparison}
\end{figure}

Meanwhile, various benchmarks have been developed to assess LLM capabilities, focusing on specific skills \citep{DBLP:conf/nips/HendrycksBKABTS21} or comprehensive abilities \citep{DBLP:conf/coling/XuHZLCLXSYYTDLS20, DBLP:conf/iclr/HendrycksBBZMSS21}. Based on these benchmarks, certain models claim to surpass larger counterparts, despite being significantly smaller in size—for instance, 7B or 13B models \citep{liu2024chatqasurpassinggpt4conversational, lei2024autocoderenhancingcodelarge} compared to GPT models.

Concerns about LLM evaluation reliability are increasing due to a lack of transparency, especially with closed-source models. It raises a crucial question: \textit{Is their impressive performance a result of generalization, or merely due to prior exposure and memorization of test samples, a problem known as data contamination?} Without effective detection methods, model providers might maliciously contaminate models to boost performance.

Data contamination detection (DCD) methods are divided into two main categories \citep{xu2024benchmarkdatacontaminationlarge}. The first is matching-based, which examines benchmark overlap \citep{DBLP:journals/corr/abs-2311-04850} or investigates whether models memorize specific samples \citep{golchin2024datacontaminationquiztool, DBLP:conf/naacl/Deng0TGC24}. A key limitation of these methods is their vulnerability to surface-level paraphrasing \citep{dekoninck2024evadingdatacontaminationdetection}. The second involves comparison-based techniques, which are more robust but often require specific conditions \citep{xu2024benchmarkingbenchmarkleakagelarge, roberts2023datacontaminationlenstime, DBLP:conf/acl/DongJLJGYL24}. In this study, we identify two main limitations in current DCD methods: (1) Existing methods do not explicitly address contamination detection in domain-specific models. As fine-tuning for specific domains becomes more common, DCD methods must distinguish between malicious contamination and legitimate fine-tuning. (2) Most DCD techniques are task-specific. For example, \citet{golchin2024datacontaminationquiztool} and \citet{DBLP:conf/naacl/Deng0TGC24} focus on text generation tasks. Extending these methods to handle a broader range of tasks remains a challenge.

In this work, we introduce a novel framework, Data Contamination Detection via \textbf{\underline{C}onsistency \underline{A}m\underline{p}lification} (\textbf{CAP}), which leverages the concept of LM consistency \citep{jang2022becel}. CAP calculates the Performance Consistency Ratio (PCR) as an indicator and compares the PCR between the training and test sets. A detailed heatmap illustrating this comparison is shown in Figure~\ref{fig:heatmap_and_comparison} (a). The direction of the difference from the training set to the test set indicates either fine-tuning or contamination. Additionally, Figure~\ref{fig:heatmap_and_comparison} (b) demonstrates that CAP has broader applicability across various tasks and model types compared to other methods.

We conducted experiments with 7 LLMs on 4 financial benchmarks, including 4 general-purpose models and 3 financial-specific LLMs (Fin-LLMs). Results show that CAP can effectively distinguish between fine-tuning and benchmark contamination. The Fin-LLMs tested are predominantly fine-tuned without intentional contamination, but we identified two key issues: (1) contamination in general-purpose models can propagate to domain-specific models fine-tuned on them, and (2) Benchmarks from public sources, like exams, textbooks, or older datasets, are prone to unintentional overlap with other data. Our contributions are threefold:
\begin{itemize}
    \item We introduce CAP, a novel method for data contamination detection, and define PCR, mathematically demonstrating that PCR increases when datasets are used in training.
    \item We are the first to explicitly investigate contamination in domain-specific LLMs, providing a comprehensive empirical evaluation of domain-specific benchmarks and models.
    \item We highlight the risks of unintentional contamination, particularly in composite benchmarks, given the rapid growth of new benchmarks.
\end{itemize}
\section{Related Work}
\begin{figure*}[ht]
    \centering
    \includegraphics[width=\textwidth]{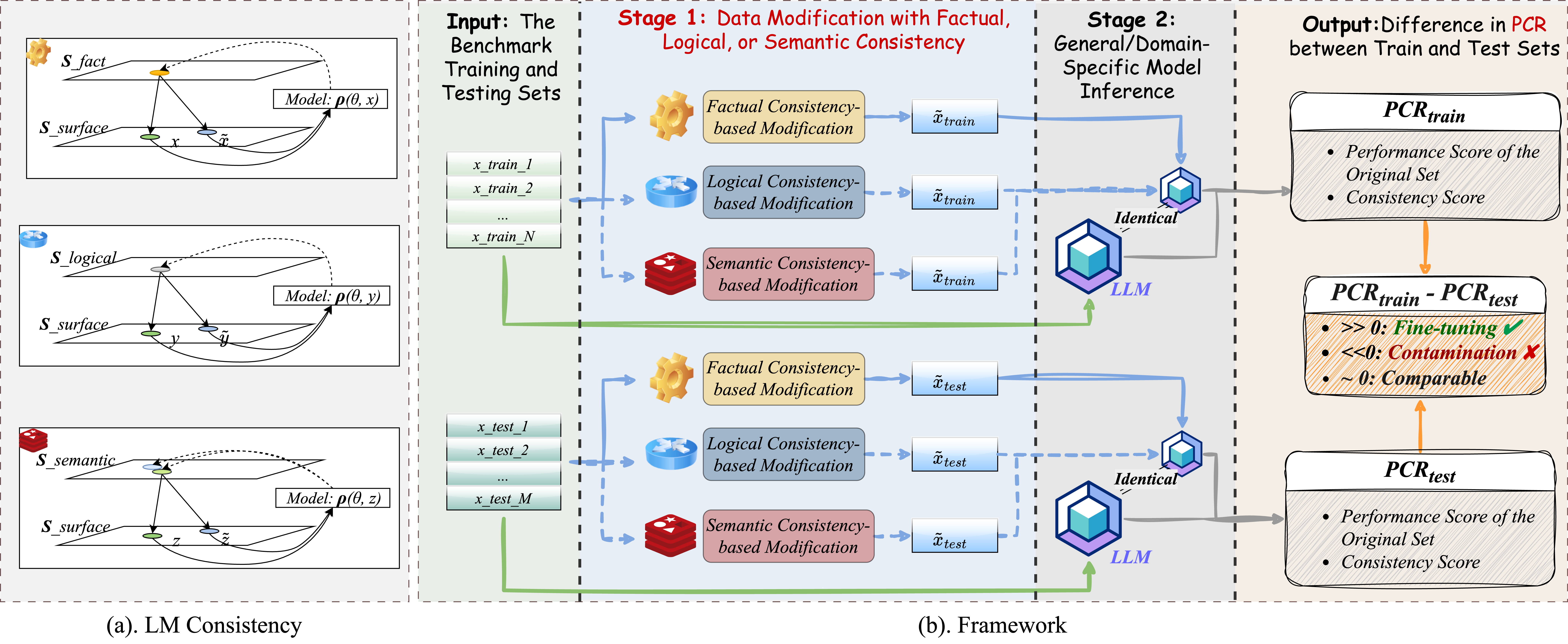}
    \caption{(a) LM Consistency: Different points in surface space may correspond to the same points in logical and factual space, or to nearby points in semantic space. (b) The CAP Framework: (i) The input stage consists of the training and test sets, covering benchmarks like multiple-choice questions (MCQ) and sequence generation tasks; (ii) Data modification is applied with a specific consistency criterion; (iii) Both the original and modified datasets are processed through LLMs, which can be black-box models; (iv) Finally, the Performance Consistency Ratio (PCR) is calculated and compared between the training and test sets to assess the presence of fine-tuning or contamination.}
    \label{fig:framework}
\end{figure*}
\subsection{Data Contamination Detection (DCD)}
 \paragraph{Matching-based DCD} Efforts to detect data contamination in LLMs can generally be categorized into two main approaches. The first is matching-based, which evaluates whether the model recalls specific training samples. Notable methods include DCQ \citep{golchin2024datacontaminationquiztool}, where the model selects the correct sample from a set that includes the original and three perturbed variants. Similarly, \citet{DBLP:conf/naacl/Deng0TGC24} introduced Testset Slot Guessing, which hides parts of the test set and prompts the LLM to predict the missing elements. Additionally, \citet{DBLP:journals/corr/abs-2311-04850} utilize similarity search to retrieve the top-k samples from a dataset, followed by prompting a strong LLM like GPT-4 to assess whether any of the top-k samples are too similar to the test case. One limitation of these methods is their vulnerability to surface-level rephrasing, where minor alterations can lead to failure  \citep{dekoninck2024evadingdatacontaminationdetection}. Furthermore, while these techniques are effective for tasks like text generation, where the outputs consist of long sequences, they may be less suitable for benchmarks like Multiple Choice Questions (MCQs), which have become increasingly popular nowadays. Our method can handle both task types effectively.
\paragraph{Comparison-based DCD} 
The second type is comparison-based. For example, CDD \citep{Dong2024GeneralizationOM} relies on the output distribution of LLMs to detect data contamination. However, it requires an ideal reference output distribution, which can be challenging to define or obtain, especially when benchmarks involve composite tasks. \citet{xu2024benchmarkingbenchmarkleakagelarge} applied perplexity for contamination detection, but their approach requires access to hidden layers during decoding, making it suitable only for gray-box or white-box models. In contrast, our method works for both types of LMs. \citet{roberts2023datacontaminationlenstime} conducted a longitudinal analysis of data contamination in LLMs before and after specific training cutoffs. However, the absence of temporal markers in certain datasets limits the application of such chronological analyses. Our method, by contrast, leverages the intrinsic consistency of LMs, reducing reliance on external resources like temporal information, making it more broadly applicable.

\subsection{Consistency in LLMs}
The concept of consistency in large language models (LLMs) was introduced by \citet{jang2022becel}, who developed a framework based on behavioral consistency and unified previous studies into a taxonomy. This framework defines consistency through two key components: \textit{belief}, referring to what the model assumes to be true, and \textit{principle}, which governs coherent decision-making. From these foundations, three main categories of consistency emerged: \textit{semantic}, \textit{logical}, and \textit{factual}.

\textbf{Semantic consistency} refers to the ability of an LLM to make the same decisions when faced with semantically equivalent texts. It aligns with Meaning-Text Theory (MTT), which posits a many-to-many relationship between linguistic forms and semantic content \citep{mel1970towards}.

\textbf{Logical consistency} refers to a model’s ability to make reasoning decisions without contradictions. It ensures that superficial changes in phrasing do not affect the model's reasoning process or outcomes. Therefore, the model should consistently generate correct predictions even when the input undergoes minor modifications.

\textbf{Factual consistency} refers to a model's ability to produce outputs that align with established facts and the provided context. This type of consistency is crucial for tasks involving knowledge grounding and utilization, as the model must avoid generating outputs that contradict the knowledge it holds.
\section{Methodology}
\begin{figure*}[htbp]
    \centering
    \includegraphics[width=\textwidth]{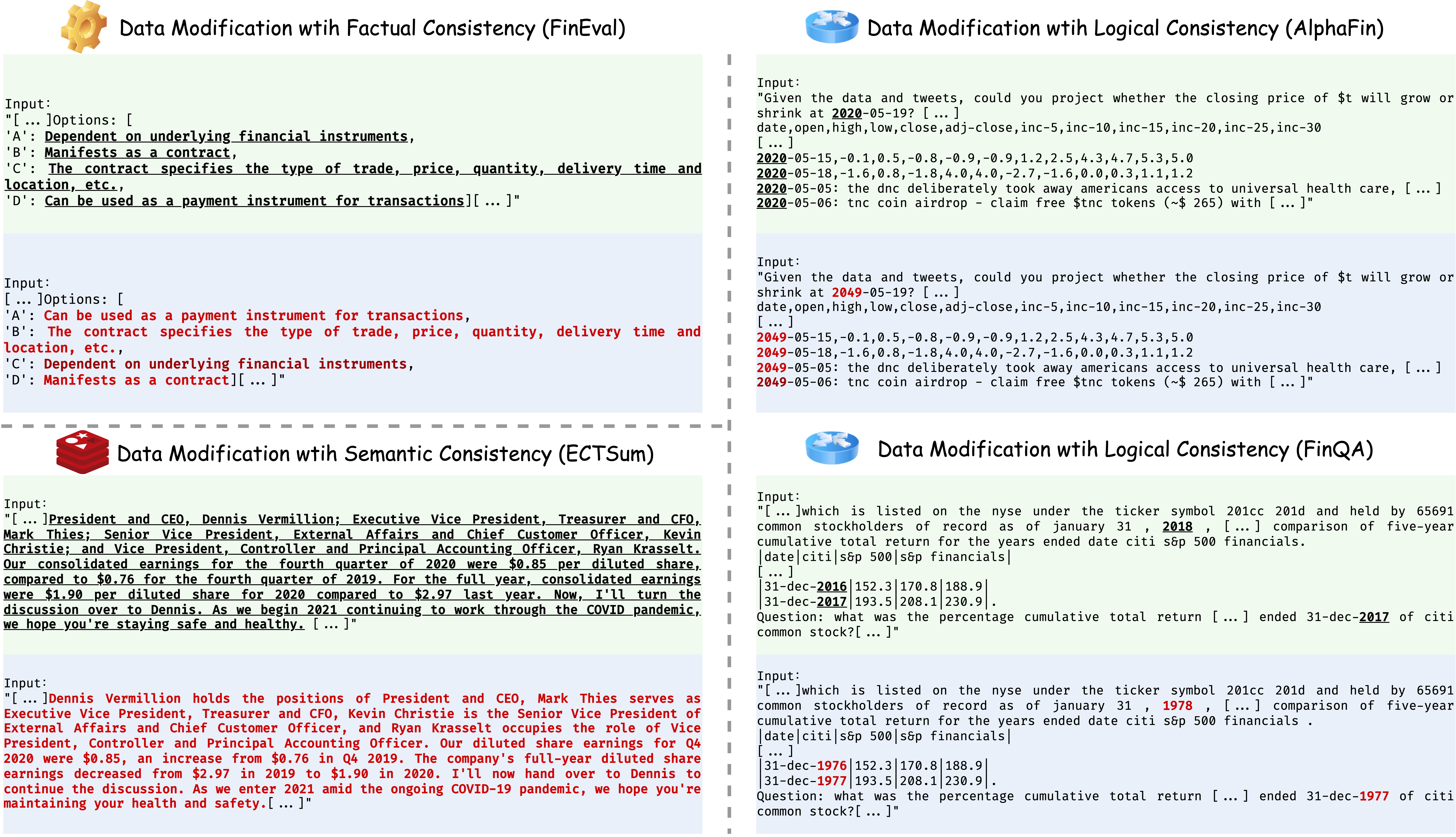}
    \caption{Consistency-based Data Modification: (1) For MCQ benchmarks like FinEval, we apply factual consistency by rearranging the corresponding content of options. (2) For Q\&A tasks like AlphaFin and FinQA, we modify logically unrelated information (e.g., `year') while keeping the reasoning process intact. (3) For summarization tasks like ECTSum, we modify the input text while maintaining semantic meaning (cosine similarity $> 0.979$).}
    \label{fig:data_modification}
\end{figure*}
\subsection{Problem Formulation}
\citet{dekoninck2024evadingdatacontaminationdetection} define benchmark-level dataset contamination as occurring when a model’s training corpus $D$ overlaps with a benchmark set $D'$. Our study refines this definition in the context of domain-specific models. We clarify that fine-tuning using the training set of a benchmark is permissible. Benchmark-level data contamination occurs when the \textbf{test set} of a benchmark is included during the pre-training or fine-tuning of a LLM. We provide the following formal definition:

\textbf{Definition (Benchmark Contamination)}: A model is considered contaminated by a benchmark if its training corpus $D$ overlaps with the benchmark's test set $D'\_test$, i.e., $D'\_test \cap D \neq \emptyset$.

When a dataset is not explicitly identified as part of a benchmark’s training or test set, we use the term \textbf{\textit{dataset leakage}} to indicate the extent to which the model has been exposed to, memorized, or trained on the dataset \citep{xu2024benchmarkingbenchmarkleakagelarge}. For domain-specific LLMs, the objective of contamination detection is to assess whether the model has been fine-tuned using the benchmark's training set or has been contaminated by its test set.
\subsection{CAP Framework}
Figure~\ref{fig:framework} (a) illustrates the concept of LM consistency. For data points $x$ and $\Tilde{x}$ in the surface space, which represent different expressions of the same fact, factual consistency encourages the model's predictions to reflect the same underlying fact or knowledge. The same principle applies to logical consistency. In the semantic space, when two surface-level data points, though textually different, are semantically equivalent, the model's outputs should exhibit a high degree of semantic similarity.

Building on these principles, we propose the \textbf{CAP} framework: Data Contamination Detection via \underline{C}onsistency \underline{A}m\underline{p}lification, as illustrated in Figure~\ref{fig:framework} (b). CAP operates through four stages: (1) At the Input Stage, two sets are provided for a specific benchmark: the training set and the test set. (2) Stage 1: Modified datasets are constructed based on factual, logical, or semantic consistency, depending on the task. For example, factual consistency is recommended for MCQs, logical consistency for reasoning tasks, and semantic consistency for tasks like sequence generation or summarization. (3) Stage 2: LM inference is performed using white-box, gray-box, or black-box models. The model is run on both the original and modified training and test sets. (4) Output Stage: The Performance Consistency Ratio (PCR) is computed for both sets. If $PCR\_train >> PCR\_test$, this suggests fine-tuning. Conversely, if $PCR\_train << PCR\_test$, it indicates test data contamination. When the difference between $PCR\_Train$ and $PCR\_test$ is close to zero, it implies no leakage or similar contamination levels in both sets.
\subsection{Performance Consistency Ratio}
This section explains the mathematical relationship between the Performance Consistency Ratio (PCR) and dataset leakage, demonstrating how PCR can be used as a measure of data contamination.  Our primary objective is to compare the degree of leakage between the training and test sets of a benchmark for a LM. Let $\rho(\cdot)$ represent a language model and $X$ an arbitrary dataset. We define $M$ as the metric score of the model's predictions, $\gamma(\cdot)$ as a consistency-based data modification operation, and $C$ as the consistency score between the model's predictions on the original and modified datasets. Suppose $X_1$ is a dataset included in the pre-training or fine-tuning set of model $\rho$. The following holds:
\begin{itemize}
\item The performance metric score $M \geq 0$ and the consistency score $C \geq 0$.
\item Generally, the performance metric of a fine-tuned dataset $M(\rho(X_1))$ is greater than or equal to $M(\rho(X))$ \citep{xu2024benchmarkingbenchmarkleakagelarge}:
\begin{equation}
\label{eq_M}
M(\rho(X_1)) \geq M(\rho(X))
\end{equation}
\item The consistency between the model outputs on datasets $X$ and $X_1$, denoted by $C(\rho(X), \rho(X_1))$, is less than or equal to the consistency between the outputs on the original dataset $X$ and its consistency-based modified version $\gamma(X)$ \citep{jang2022becel}:
\begin{equation}
\label{eq_C1}
C(\rho(X), \rho(X_1)) \leq C(\rho(X), \rho(\gamma(X)))
\end{equation}
\end{itemize}
From Inequality~(\ref{eq_C1}), by substituting $X = X_1$ on the right-hand side, we obtain:
\begin{equation}
\label{eq_C2}
C(\rho(X), \rho(X_1)) \leq C( \rho(X_1), \rho(\gamma(X_1)))
\end{equation}
Considering Inequalities~(\ref{eq_M}) and (\ref{eq_C1}), we can express the following:
\begin{equation}
\label{eq_R1}
\frac{M(\rho(X_1))}{C(\rho(X), \rho(X_1))} \geq \frac{M(\rho(X))}{C(\rho(X), \rho(\gamma(X)))}
\end{equation}
Furthermore, considering Inequality~(\ref{eq_C2}), for the left-hand side of Inequality (\ref{eq_R1}), we obtain:
\begin{equation}
\label{eq_R2}
\frac{M(\rho(X_1))}{C(\rho(X_1), \rho(\gamma(X_1)))} \geq \frac{M(\rho(X_1))}{C(\rho(X), \rho(X_1))}  
\end{equation}
By combining (\ref{eq_R1}) and (\ref{eq_R2}), we derive:
\begin{equation}
\label{eq_R3}
\frac{M(\rho(X_1))}{C(\rho(X_1), \rho(\gamma(X_1)))} \geq \frac{M(\rho(X))}{C(\rho(X), \rho(\gamma(X)))}
\end{equation}
We define the Performance Consistency Ratio (\textbf{PCR}) as:
\begin{equation}
PCR=PCR (X, \gamma, \rho )=\frac{M(\rho(X))}{C(\rho(X), \rho(\gamma(X)))}
\end{equation}
From Inequality~(\ref{eq_R3}), it follows that the PCR for a dataset included in the pre-training or fine-tuning set of the language model will be larger.
\subsection{Practical Implementation of the Performance Consistency Ratio}
In practical applications, the consistency metric $C$ can take on very small values, and in some cases, it may approach zero. To mitigate the risk of division by zero, we introduce a negligibly small constant $\alpha$ during the computation. The adjusted Performance Consistency Ratio (PCR) is therefore defined as:
\begin{equation}
PCR = \frac{M(\rho(X)) + \alpha}{C(\rho(X), \rho(\gamma(X))) + \alpha} 
\end{equation}
Moreover, to ensure that the resulting values are within a suitable range, we apply a $Tanh(\cdot)$ transformation. The practical formula for the PCR is:
\begin{equation}
PCR = \mathit{Tanh}\left(\frac{M(\rho(X)) + \alpha}{C(\rho(X), \rho(\gamma(X))) + \alpha}\right)
\end{equation}
\subsection{Consistency-based Data Modification}
This section introduces the consistency-based data modification operation, $\gamma(\cdot)$, for a dataset $X$. As depicted in Figure~\ref{fig:framework}, there are three types of consistency: factual, logical, and semantic. To evaluate the consistency metric before and after data modification, it is essential that the modification operation aligns with the specified type of consistency during the synthesis of the modified data. Examples of each consistency type are provided in Figure~\ref{fig:data_modification}.
\paragraph{Benchmark Collection} 
Our research focuses on methods for detecting dataset contamination in both general-purpose and domain-specific models. Given the recent rapid growth of the financial language models (Fin-LLMs), this study focuses on this domain. Key open-source benchmarks include FinQA \citep{DBLP:conf/emnlp/ChenCSSBLMBHRW21}, ECTSum \citep{DBLP:conf/emnlp/MukherjeeBBSHSS22}, FinEval \citep{Zhang2023FinEvalAC} and AlphaFin \citep{Li2024AlphaFinBF}, among others.

\paragraph{Factual-Consistency-Based Data Modification} FinEval consists of MCQs designed to assess the model's ability to leverage financial knowledge. For this benchmark, we measure the factual consistency of the model's predictions before and after applying modifications. The modification involves rearranging the content of choices `A', `B', `C', and `D'. An example is provided in Figure~\ref{fig:data_modification}. 

\paragraph{Logical-Consistency-Based Data Modification} 
FinQA is a benchmark for table-based and textual question answering (Q\&A), designed to assess models' reasoning and calculation abilities. For this benchmark, we focus on the logical consistency of the model's answers before and after applying modifications. Similarly, AlphaFin, a composite benchmark, includes both traditional financial question-answering tasks and more advanced tasks such as stock prediction. Our focus is on the Q\&A component of AlphaFin. For these two benchmarks, we use a minor modification approach by changing the `year' information in the data, while the logical calculation should yield the same result. Examples are provided in Figure~\ref{fig:data_modification}.

\paragraph{Semantic-Consistency-Based Data Modification} 
ECTSum is a summarization benchmark that evaluates the sequence generation capabilities of LMs. We measure the semantic consistency of output summaries before and after applying modifications to the input. Model \texttt{claude-3.5-sonnet-20240620} is used to construct semantically similar versions of the original long text. The prompt is in Table~\ref{tab:prompt_ectsum} in Appendix~\ref{sec:appendix_prompt}, and an example is shown in Figure~\ref{fig:data_modification}. 
\section{Experiment Settings}
\begin{table}[]
\centering
\resizebox{\columnwidth}{!}{%
\begin{tabular}{cc|ccc|cc|ccc}
\hline
\multicolumn{2}{c|}{\textbf{FinEval}} & \multicolumn{3}{c|}{\textbf{FinQA}} & \multicolumn{2}{c|}{\textbf{AlphaFin}} & \multicolumn{3}{c}{\textbf{ECTSum}} \\ \hline
\#Dev & \#Val & \#Train & \#Val & \#Test & \#Train & \#Val & \#Train & \#Val & \#Test \\ \hline
\multirow{2}{*}{170} & \multirow{2}{*}{1151} & 500 & 500 & 500 & \multirow{2}{*}{500} & \multirow{2}{*}{500} & 517 & 101 & 250 \\
 &  & 100\textsuperscript{+} & 100\textsuperscript{+} & 100\textsuperscript{+} &  &  & 100\textsuperscript{*} & 100\textsuperscript{*} & 100\textsuperscript{*} \\ \hline
\end{tabular}%
}
\caption{Dataset size statistics. In FinQA, `+’ indicates Baichuan and Disc-Fin used 100 samples. In ECTSum, `*’ denotes LLaMA and Baichuan used 100 samples.}
\label{tab:data_size}
\end{table}
\begin{figure*}[htbp]
    \centering
        \begin{subfigure}{0.245\textwidth}
        \includegraphics[width=\linewidth]{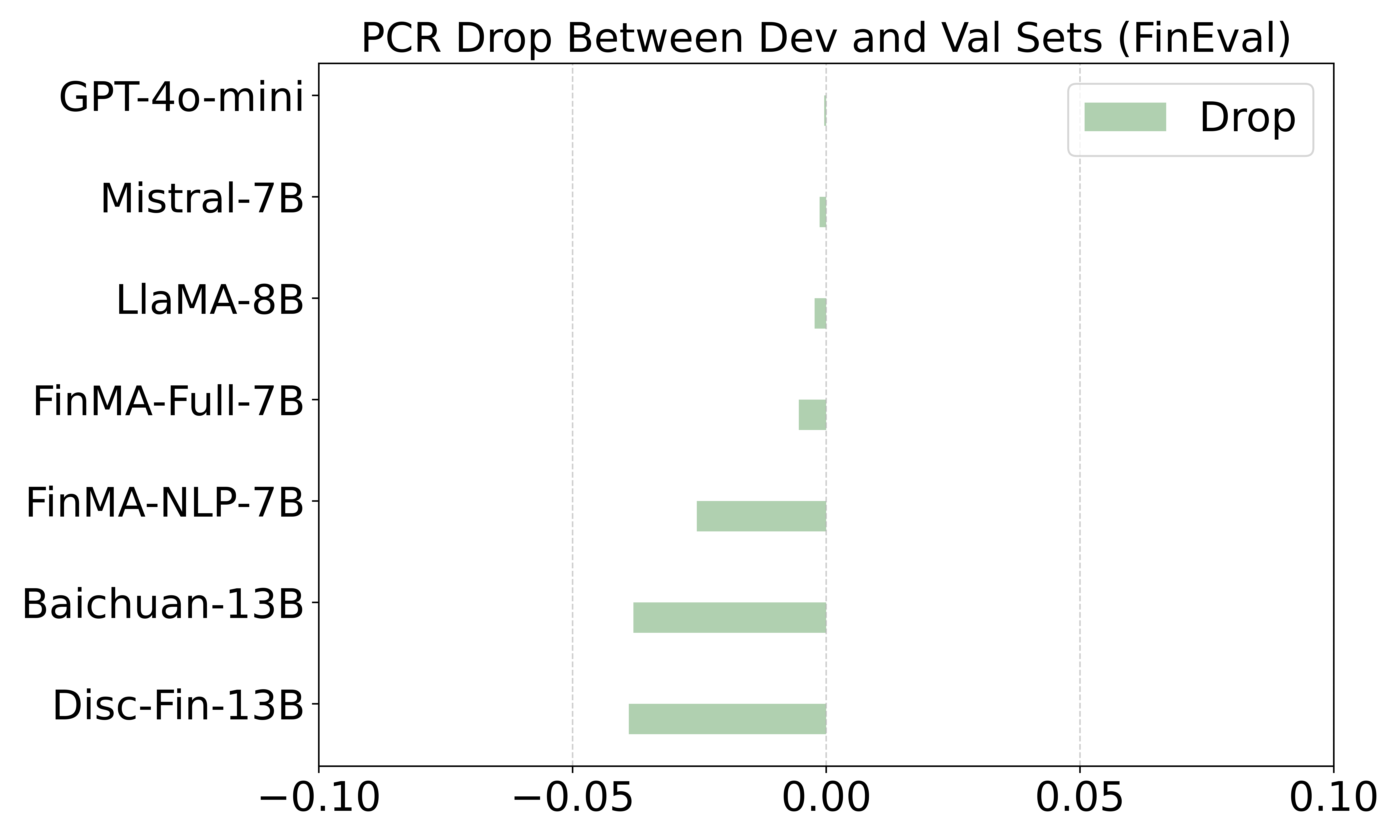}
        \caption{FinEval}
    \end{subfigure}
    \begin{subfigure}{0.245\textwidth}
        \includegraphics[width=\linewidth]{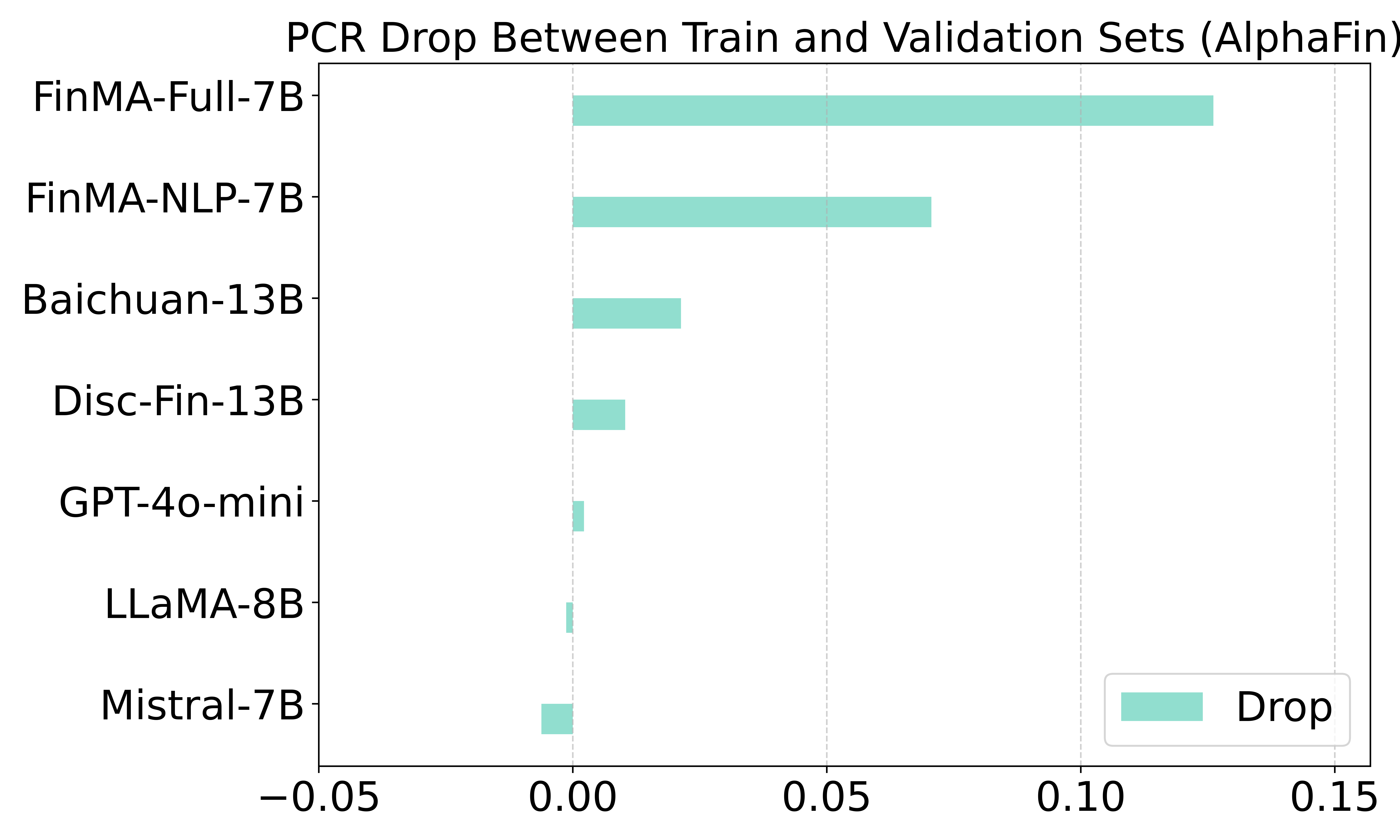}
        \caption{AlphaFin Research}
    \end{subfigure}
    \begin{subfigure}{0.245\textwidth}
        \includegraphics[width=\linewidth]{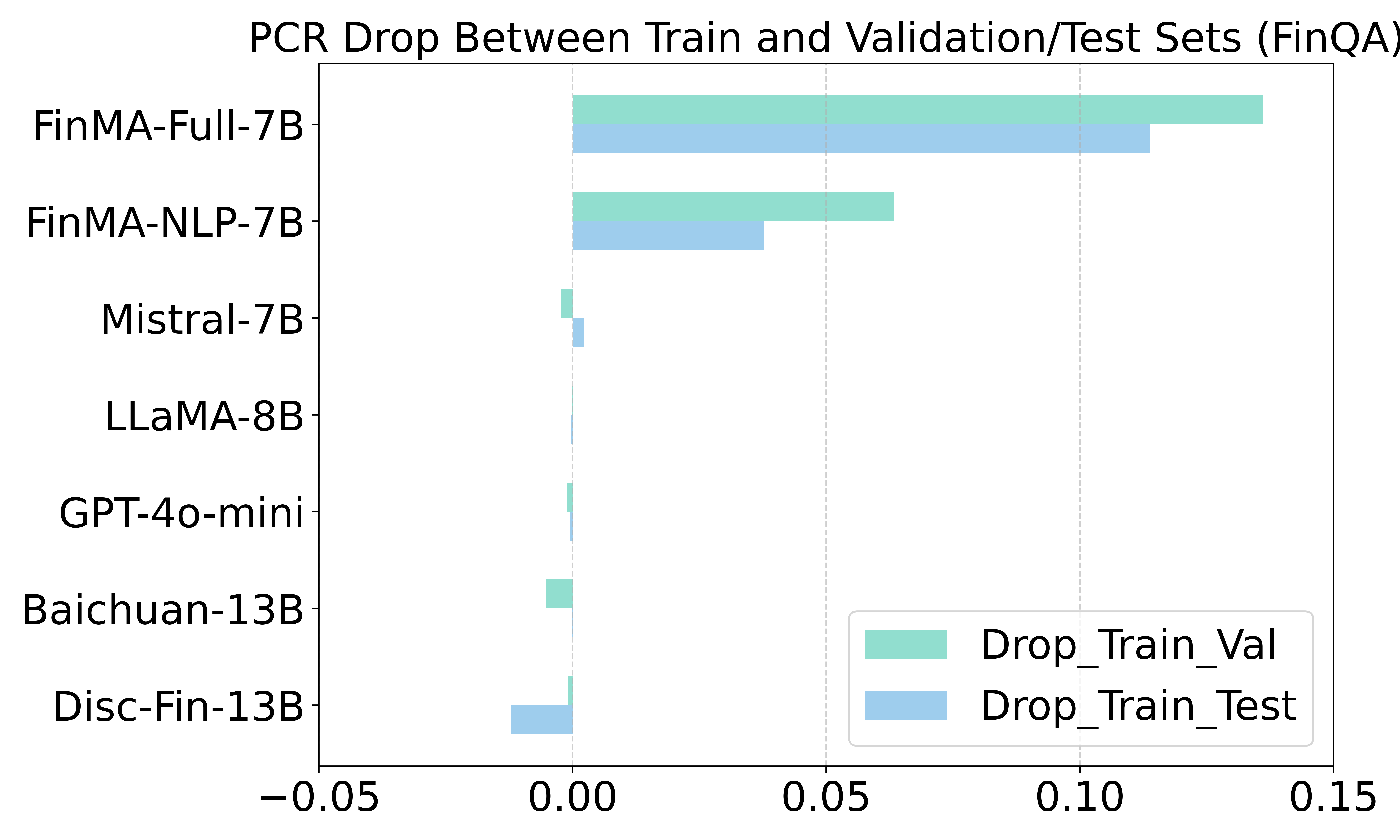}
        \caption{FinQA}
    \end{subfigure}
    \begin{subfigure}{0.245\textwidth}
        \includegraphics[width=\linewidth]{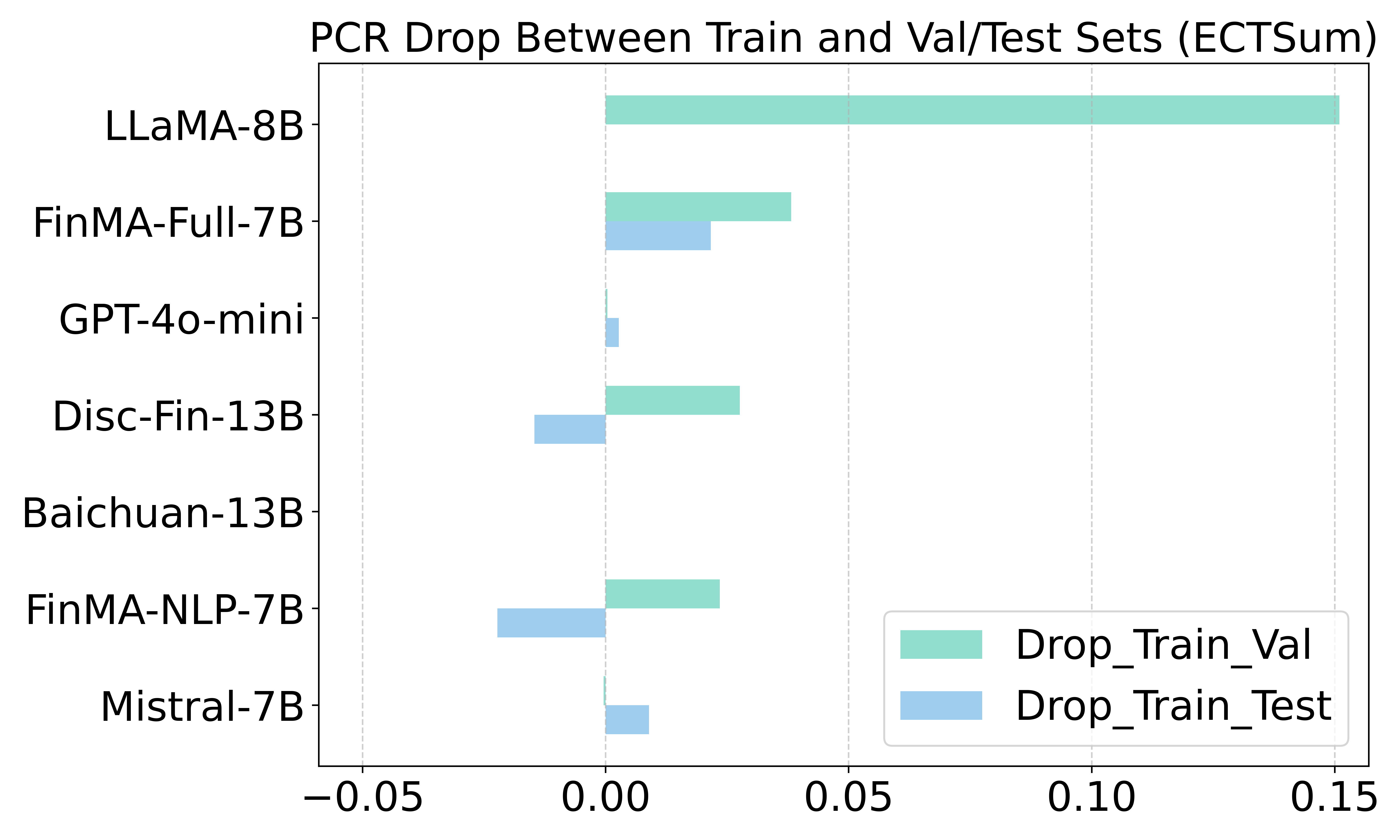}
        \caption{ECTSum}
    \end{subfigure}
    \caption{ Illustration of Four Benchmarks: (a) Baichuan and Disc-Fin may have been contaminated by the FinEval validation set; (b) and (c) FinMA-Full and FinMA-NLP were fine-tuned with the FinQA training set. AlphaFin Research, though newly proposed, has overlap with FinQA; (d) ECTSum is less likely to be used, with most values close to zero. However, it suggests LLaMa may have been contaminated by the training and test sets.}
    \label{fig:res_main}
\end{figure*}
\begin{table*}[t]
\centering
\resizebox{\textwidth}{!}{%
\begin{tabular}{lrrrrrrrrrrr}
\toprule
 & \multicolumn{4}{c}{\textbf{FinEval}} & \multicolumn{7}{c}{\textbf{FinQA}} \\ 
\cmidrule(lr){2-5} \cmidrule(lr){6-12}
 & \multicolumn{3}{c}{\textbf{Baselines}} & \multicolumn{1}{c}{\multirow{2}{*}{\begin{tabular}[c]{@{}c@{}}\textbf{CAP}\\ (Ours)\end{tabular}}} & \multicolumn{5}{c}{\textbf{Baselines}} & \multicolumn{2}{c}{\multirow{2}{*}{\begin{tabular}[c]{@{}c@{}}\textbf{CAP}\\ (Ours)\end{tabular}}} \\ 
\cmidrule(lr){2-4} \cmidrule(lr){6-10}
 & \multicolumn{2}{c}{\textbf{Performance Decline ($\Delta$)}} & \multicolumn{1}{c}{\textbf{Discrepancy Comparison ($\delta$)}} & & \multicolumn{3}{c}{\textbf{Performance Decline ($\Delta$)}} & \multicolumn{2}{c}{\textbf{Discrepancy Comparison ($\delta$)}} & & \\ 
\cmidrule(lr){2-3}\cmidrule(lr){6-8}\cmidrule(lr){9-10}\cmidrule(lr){11-12}
Model & \multicolumn{1}{c}{$\Delta_{Dev}$} & \multicolumn{1}{c}{$\Delta_{Val}$} & \multicolumn{1}{c}{$\delta_{\text{dev-val}}$} & \multicolumn{1}{c}{\textit{PCR}$_{\text{dev-val}}$} & \multicolumn{1}{c}{$\Delta_{Train}$} & \multicolumn{1}{c}{$\Delta_{Val}$} & \multicolumn{1}{c}{$\Delta_{Test}$} & \multicolumn{1}{c}{$\delta_{\text{train-val}}$} & \multicolumn{1}{c}{$\delta_{\text{train-test}}$} & \multicolumn{1}{c}{\textit{PCR}$_{\text{train-val}}$} & \multicolumn{1}{c}{\textit{PCR}$_{\text{train-test}}$} \\ 
\midrule
LLaMA-8B & 0 & 0 & 0 & -0.0023 & -0.0001 & -0.0001 & -0.0002 & 0 & 0.0433 & -0.0001 & -0.0003 \\
Mistral-7B & 0 & 0 & 0 & -0.0013 & -0.0011 & 0.0004 & -0.0004 & -0.0701 & -0.0327 & -0.0023 & 0.0023 \\
FinMA-Full-7B & \underline{0.0471} & 0.0059 & \underline{0.2235} & -0.0054 & \textbf{0.0123} & \textbf{0.0108} & \textbf{0.0068} & -0.0611 & -0.0093 & \textbf{0.136} & \textbf{0.1139} \\
FinMA-NLP-7B & \underline{-0.0471} & \underline{-0.0417} & -0.0197 & -0.0255 & \underline{0.009} & \underline{0.0054} & 0 & 0.0021 & 0.0608 & \underline{0.0633} & \underline{0.0377} \\
Baichuan-13B & -0.0177 & 0.0235 & -0.0935 & \underline{-0.038} & 0.0006 & -0.0031 & 0.0013 & \textbf{0.3814} & \underline{-0.0891} & -0.0053 & -0.0001 \\
Disc-Fin-13B & \textbf{0.0588} & \textbf{0.0443} & 0.0241 & \textbf{-0.0389} & 0.0007 & -0.0017 & \underline{-0.0015} & \underline{0.2162} & \textbf{0.1763} & -0.0009 & -0.0121 \\
GPT-4o-mini & 0 & -0.0009 & \textbf{9} & -0.0004 & -0.0008 & 0.0011 & 1e-04 & -0.089 & -0.0431 & -0.001 & -0.0005 \\
\midrule
 & \multicolumn{4}{c}{\textbf{AlphaFin}} & \multicolumn{7}{c}{\textbf{ECTSum}} \\ 
\cmidrule(lr){2-5} \cmidrule(lr){6-12}
 & \multicolumn{3}{c}{\textbf{Baselines}} & \multicolumn{1}{c}{\multirow{2}{*}{\begin{tabular}[c]{@{}c@{}}\textbf{CAP}\\ (Ours)\end{tabular}}} & \multicolumn{5}{c}{\textbf{Baselines}} & \multicolumn{2}{c}{\multirow{2}{*}{\begin{tabular}[c]{@{}c@{}}\textbf{CAP}\\ (Ours)\end{tabular}}} \\ 
\cmidrule(lr){2-4} \cmidrule(lr){6-10}
 & \multicolumn{2}{c}{\textbf{Performance Decline ($\Delta$)}} & \multicolumn{1}{c}{\textbf{Discrepancy Comparison ($\delta$)}} & & \multicolumn{3}{c}{\textbf{Performance Decline ($\Delta$)}} & \multicolumn{2}{c}{\textbf{Discrepancy Comparison ($\delta$)}} & & \\ 
\cmidrule(lr){2-3}\cmidrule(lr){6-8}\cmidrule(lr){9-10}\cmidrule(lr){11-12}
Model & \multicolumn{1}{c}{$\Delta_{Train}$} & \multicolumn{1}{c}{$\Delta_{Val}$} & \multicolumn{1}{c}{$\delta_{\text{train-val}}$} & \multicolumn{1}{c}{\textit{PCR}$_{\text{train-val}}$} & \multicolumn{1}{c}{$\Delta_{Train}$} & \multicolumn{1}{c}{$\Delta_{Val}$} & \multicolumn{1}{c}{$\Delta_{Test}$} & \multicolumn{1}{c}{$\delta_{\text{train-val}}$} & \multicolumn{1}{c}{$\delta_{\text{train-test}}$} & \multicolumn{1}{c}{\textit{PCR}$_{\text{train-val}}$} & \multicolumn{1}{c}{\textit{PCR}$_{\text{train-test}}$} \\ 
\midrule
LLaMA-8B & -1e-04 & 0.0005 & \textbf{-0.0651} & -0.0013 & 0.0024 & \underline{0.0024} & \underline{0.0011} & -0.0095 & 0.0356 & \textbf{0.1509} & 0 \\
Mistral-7B & -0.0015 & -0.003 & 0.0121 & -0.0062 & \textbf{0.0052} & \textbf{0.0032} & \textbf{0.0032} & 0.0388 & 0.0332 & -0.0004 & 0.0089 \\
FinMA-Full-7B & \textbf{0.034} & -0.0047 & \underline{0.0518} & \textbf{0.1261} & 1e-04 & -0.0005 & 0.0002 & 0.2917 & -0.0453 & \underline{0.0382} & \underline{0.0216} \\
FinMA-NLP-7B & 0.0033 & 0.0014 & 0.0031 & \underline{0.0706} & -0.0002 & -0.0007 & -0.0002 & \textbf{6.8462} & 0.028 & 0.0235 & \textbf{-0.0223} \\
Baichuan-13B & 0.0022 & \underline{0.0049} & -0.032 & 0.0213 & 0.0024 & 0.0017 & -0.001 & 0.1285 & \textbf{0.5855} & 0 & 0 \\
Disc-Fin-13B & 0.0006 & 0.0009 & -0.0045 & 0.0103 & -0.0015 & -0.0004 & 0.0008 & \underline{-0.2771} & \underline{-0.5402} & 0.0276 & -0.0147 \\
GPT-4o-mini & \underline{-0.0046} & \textbf{-0.0144} & 0.0383 & 0.0022 & \underline{0.0032} & -0.0011 & 0.001 & 0.0926 & 0.0447 & 0.0004 & 0.0027 \\
\bottomrule
\end{tabular}%
}
\caption{Comparison of results across the FinEval, FinQA, AlphaFin, and ECTSum benchmarks for various models using Baseline and CAP (Ours) methods. The largest absolute values are in \textbf{bold}, and the second largest are \underline{underlined}. FinMA-Full and FinMA-NLP are fine-tuned with the FinQA training set, aligning with CAP's findings \citep{DBLP:conf/nips/XieHZLPLH23}. FinEval shows overlap with Baichuan’s fine-tuning set, further supporting CAP's results \citep{baichuan2023baichuan}. There is insufficient evidence to determine whether ECTSum has contaminated models.}
\label{tab:res_main}
\end{table*}

\begin{table*}[t]
    \centering
    \resizebox{\textwidth}{!}{%
    \begin{tabular}{lcccccccccccc}
    \toprule
    & \multicolumn{4}{c}{$PCR_{dev-val}$ with $\tanh(\cdot)$} & \multicolumn{4}{c}{$PCR_{dev-val}$ w/o $\tanh(\cdot)$} & \multicolumn{4}{c}{$PCR_{dev-val}$ with $\sigma (\cdot)$} \\
    \cmidrule(lr){2-5} \cmidrule(lr){6-9} \cmidrule(lr){10-13}
    Model & $\alpha=0$ & $\alpha=0.001$ & $\alpha=0.01$ & $\alpha=0.1$ & $\alpha=0$ & $\alpha=0.001$ & $\alpha=0.01$ & $\alpha=0.1$ & $\alpha=0$ & $\alpha=0.001$ & $\alpha=0.01$ & $\alpha=0.1$ \\
    \midrule
    LLaMA-8B      & 0       & -0.0002 & -0.0023 & -0.0165 & 0       & -0.0002 & -0.0023 & -0.0169 & 0       & -0.0001 & -0.0006 & -0.0042 \\
    Mistral-7B    & 0       & -0.0001 & -0.0013 & -0.0091 & 0       & -0.0001 & -0.0013 & -0.0094 & 0       & 0       & -0.0003 & -0.0023 \\
    FinMA-Full-7B & -0.0051 & -0.0051 & -0.0054 & -0.0062 & -0.0078 & -0.0079 & -0.0084 & -0.0107 & -0.0017 & -0.0018 & -0.0019 & -0.0023 \\
    FinMA-NLP-7B  & -0.0245 & -0.0246 & -0.0255 & -0.0264 & -0.0296 & -0.0298 & -0.0313 & -0.037  & -0.0071 & -0.0071 & -0.0074 & -0.0085 \\
    Baichuan-13B  & -0.0389 & -0.0388 & -0.038 & -0.0316 & -0.0834 & -0.0832 & -0.0817 & -0.069  & -0.0169 & -0.0169 & -0.0166 & -0.0139 \\
    Disc-Fin-13B  & -0.0398 & -0.0397 & -0.0389 & -0.0325 & -0.0965 & -0.0962 & -0.0943 & -0.0784 & -0.0189 & -0.0188 & -0.0184 & -0.0153 \\
    GPT-4o-mini   & 0       & 0       & -0.0004 & -0.0029 & 0       & 0       & -0.0004 & -0.0029 & 0       & 0       & -0.0001 & -0.0007 \\
    \bottomrule
    \end{tabular}%
    }
    \caption{Performance Consistency Ratios (PCR) under Different Scale Functions and $\alpha$ Values: (1) While the choice of scale function does not significantly impact the final contamination determination, the $tanh$ function provides more suitable output ranges. (2) The value of $\alpha$ generally does not significantly impact the final determination.}
    \label{tab:res_ab}
\end{table*}

\subsection{Models \& Benchmarks \& Baselines}
\paragraph{Models} The Fin-LLMs used in our experiments include FinMA \citep{DBLP:conf/nips/XieHZLPLH23} and Disc-Fin \citep{Chen2023DISCFinLLMAC}, which have been fine-tuned on different base models: FinMA is fine-tuned on LLaMA-2 \citep{touvron2023llama2openfoundation}, while Disc-Fin is fine-tuned on Baichuan \citep{baichuan2023baichuan}. Additionally, we incorporated several widely used general-purpose LLMs to provide a comprehensive evaluation. These models include LLaMA, Mistral, Baichuan, and GPT. The specific versions are: Llama-3-8B-Instruct, Mistral-7B-Instruct-v0.3, FinMA-7B-Full, FinMA-7B-NLP, Baichuan-13B-Chat, DISC-FinLLM-13B, and \texttt{gpt-4o-mini-2024-07-18}.
\paragraph{Benchmarks and Datasets} Our experiments span a diverse range of NLP tasks, utilizing benchmarks that include FinQA, ECTSum, FinEval, and AlphaFin. Dataset sizes are shown in Table~\ref{tab:data_size}.
\paragraph{Baselines}
We compare our method to two main approaches. The first examines performance degradation as an indicator of contamination, following \citet{roberts2023datacontaminationlenstime}, \citet{DBLP:conf/acl/HuangL0GLLLS0DC24}, and \citet{DBLP:conf/aaai/LiF24}. The second measures the discrepancy between training and test performance drop, calculated as $\delta_{\text{train-test}} = \delta_{\text{train}} - \delta_{\text{test}}$, where $\delta = \Delta / M_{\text{ori}}$, $\Delta = M_{\text{ori}} - M_{\text{mod}}$, and $M$ represents metric scores, following \citet{xu2024benchmarkingbenchmarkleakagelarge}.
\subsection{Metrics}
\paragraph{Performance Metrics} 
The evaluation metrics were selected based on the characteristics of each benchmark. For FinEval, we used Exact Match (EM) \citep{DBLP:conf/acl/KamallooDCR23} as the metric. FinQA and AlphaFin were evaluated using ROUGE-L \citep{lin-2004-rouge}, which closely aligns with ROUGE-1 but provides a more comprehensive measure of answer quality. For ECTSum, given its focus on summarization, ROUGE-2 was chosen to assess the generated sequences, as it imposes stricter requirements for alignment with the reference summaries.
\paragraph{Consistency Score}
The consistency score across the three scenarios can be calculated as follows:
\begin{equation}
\left\{
\begin{array}{ll}
C_{\text{factual}} = C\left( X, \gamma_{\text{factual}} \left( X \right) \right) \\
C_{\text{logical}} = C\left( X, \gamma_{\text{logical}} \left( X \right) \right) \\
C_{\text{semantic}} = C\left( X, \gamma_{\text{semantic}} \left( X \right) \right) \\
\end{array}
\right.
\end{equation}
\begin{equation}
C\left( X, \gamma\left( X \right) \right) = \frac{1}{N}\sum_{i=1}^{N} \mathcal{I} \left( \rho\left( x_{i} \right), \rho\left( \gamma\left( x_{i} \right) \right)\right)
\end{equation}
where $\mathcal{I}$ is an indicator function that determines whether $\rho(x_{i})$ and $\rho(\gamma(x_{i}))$ are equivalent, and $x_{i} \in X = \left\{ x_{1}, \cdots, x_{N} \right\}$. The definition of $\mathcal{I}$ depends on the type of consistency:
\begin{itemize}
    \item \textbf{Factual Consistency}: $\mathcal{I}$ is set to 1 if $\rho(x_{i})$ and $\rho(\gamma(x_{i}))$ are identical; otherwise, it is 0.
    \item \textbf{Logical Consistency}: $\mathcal{I}$ is approximated by computing the Jaccard distance \citep{DBLP:conf/acl/SongLTBK24} between $\rho(x_{i})$ and $\rho(\gamma(x_{i}))$.
    \item \textbf{Semantic Consistency}: $\mathcal{I}$ is based on the cosine similarity of embedding vectors for $\rho(x_{i})$ and $\rho(\gamma(x_{i}))$, with $\mathcal{I} = 1$ if the similarity exceeds a specified threshold; otherwise, it is 0. The threshold is set to 0.95 in this study.
\end{itemize}

\section{Results and Discussion}
Table \ref{tab:res_main} summarizes the results for the seven models across the four benchmarks.
The PCR discrepancies, $PCR_{train-test}$, are calculated as $PCR_{train-test} = PCR_{train} - PCR_{test}$. The results for each benchmark are displayed in Figure~\ref{fig:res_main}.
A substantially negative $PCR_{train-test}$ value suggests potential contamination within the test set, while a significantly positive value indicates fine-tuning on the training set. A $PCR_{train-test}$ value close to zero implies comparable exposure between the training and test sets. Detailed experimental results are provided in Appendix~\ref{sec:appendix_result}.
\subsection{FinEval Results and Analysis}
We use the terms 'dev' and 'val' sets as defined in the FinEval benchmark. Notably, the test set does not provide public answers. As shown in Figure \ref{fig:res_main} (a), Baichuan-13B and Disc-Fin-13B are highly likely trained using the FinEval validation set. Baichuan-13B was fine-tuned on the C-EVAL benchmark \citep{DBLP:conf/nips/HuangBZZZSLLZLF23}, and we found overlap between C-EVAL and FinEval, with 3 identical questions present in both validation sets. These samples are provided in Appendix~\ref{sec:appendix_overlap}. Additionally, we discovered 21 overlapping samples between the FinEval validation set and the C-EVAL test set.

Although Disc-Fin-13B's instruction-tuning set does not explicitly include FinEval, it was fine-tuned on Baichuan-13B, suggesting that data contamination might be inherited. In addition, this overlap highlights the risk that benchmarks constructed from public sources, such as standardized exams (e.g., Chartered Financial Analyst [CFA]) or textbooks, may unintentionally overlap with others.
\subsection{FinQA and AlphaFin Results and Analysis}
As shown in Figure \ref{fig:res_main} (c), both FinMA-Full and FinMA-NLP were likely fine-tuned using the FinQA training set. This observation is consistent with the FinMA paper \citep{DBLP:conf/nips/XieHZLPLH23}, which notes that these models were trained on a combination of instruction datasets, including FinQA. Additionally, it confirms CAP's effectiveness in distinguishing between fine-tuning and dataset contamination, where baseline methods were insufficient, as shown in Table~\ref{tab:res_main}.

Although AlphaFin is a newer benchmark, our analysis suggests that the FinMA models may have been indirectly exposed to the AlphaFin training set. A deeper investigation into AlphaFin revealed that its QA component includes the FinQA benchmark. This raises concerns about building composite benchmarks from older datasets. LLM performance on new benchmarks may be inflated due to prior exposure to overlapping data between the older datasets and the new composite benchmark.
\subsection{ECTSum Results and Analysis}
For the ECTSum benchmark, there is insufficient evidence to verify whether it has contaminated the models. Most LLMs show $PCR_{train-val}$ and $PCR_{train-test}$ values close to zero, indicating either no contamination has occurred or both sets have been similarly contaminated. However, LLaMA shows a possible issue, with its values suggesting it may have been fine-tuned on the training set and possibly contaminated by the test set. Since the LLaMA paper \citep{touvron2023llama2openfoundation} doesn't provide detailed information on this, further investigation might be necessary.

\subsection{Ablation Studies}
We conducted ablation studies to examine the impact of different function settings and values of $\alpha$ on the performance of our method. The functions tested for scaling included $tanh$, $linear$ (w/o $tanh$), and $sigmoid$. The values of $\alpha$ explored were $0$, $0.001$, $0.01$, and $0.1$. While the results shown in Table~\ref{tab:res_ab} focus on the FinEval benchmark, similar trends were observed across other tasks. we have the following observations:
\begin{itemize}
    \item The value of $\alpha$ generally does not affect the final determination of whether a dataset is contaminated. Across the range of $\alpha$ values (0, 0.001, 0.01, and 0.1), the PCR value changes minimally. However, for tasks with smaller consistency values, such as ECTSum, we recommend using a smaller $\alpha$ (e.g., 0.001), while for other tasks, $\alpha = 0.01$ is preferable.
    \item Similarly, the choice of scale function does not significantly influence the final determination. However, using the $tanh$ function results in more appropriate output value ranges, making it the preferred choice. In contrast, $linear$ (without $tanh$) tends to produce unbounded output ranges $(-\infty, +\infty)$, while $sigmoid$ constrains the output to a narrow range $(0, 1)$. Therefore, we selected $tanh$ (which outputs within $(-1, 1)$) as the default scale function.
\end{itemize}
\section{Conclusion}
In this work, we introduced CAP, a novel framework for detecting data contamination by leveraging the Performance Consistency Ratio (PCR). CAP is versatile and can be applied across a wide range of tasks and models.
We are the first to explicitly investigate contamination in domain-specific LLMs, and our experiments across multiple LLMs and benchmarks demonstrate CAP's effectiveness in distinguishing between fine-tuning and contamination.
We also emphasize the urgent need for increased awareness among researchers and the community as new benchmarks are developed. Specifically, (1) some models may achieve inflated performance on new composite benchmarks due to pretraining or fine-tuning on older overlapping datasets; and (2) benchmarks derived from public exams or textbooks are particularly susceptible to unintentional overlap and data leakage.
%\clearpage
\section*{Limitations}
While we have demonstrated that our method effectively detects benchmark-level contamination in domain-specific LLMs, there are two main limitations that suggest promising research directions: 
\paragraph{Fin-LLMs Coverage} Due to computational constraints, we were unable to evaluate larger models such as InvestLM-65B, which is a well-known Fin-LLM. While we included FinGPT in our study, we found that its available open-source checkpoints are primarily optimized for sentiment analysis tasks, offering limited responses to financial knowledge and reasoning questions. As a result, our experiments did not include these two LLMs.

\paragraph{Contamination Focus} Furthermore, our method primarily addresses contamination at the benchmark level. It does not focus on detecting contamination at the individual sample level. Future work may explore more granular contamination detection methods at the sample level.

\section*{Ethical Considerations}
Our research focuses on public datasets and does not involve the use of sensitive data. We evaluate Fin-LLMs and detect financial benchmark contamination. These models and benchmarks are not intended for real-world financial investment decisions, and misuse could result in significant financial risks. We strongly advise against using them in such contexts without appropriate guidance.
\bibliography{custom}
\onecolumn
\appendix
\clearpage
\section{Result Appendix}
\label{sec:appendix_result}
Detailed experimental results are presented in the following tables. Table \ref{tab:fineval_results_all} provides the results for the FinEval benchmark, while Table \ref{tab:fineval_consis} reports the PCR calculation results for FinEval. Table \ref{tab:finqa_results_all} shows the results for FinQA, and Table \ref{tab:finqa_consis} presents the corresponding PCR calculations. Similarly, Table \ref{tab:alphafin_results_all} reports the results for AlphaFin, with the PCR calculations detailed in Table \ref{tab:alphafin_consis}. Finally, Table \ref{tab:ectsum_results_all} provides the results for ECTSum, and Table \ref{tab:ectsum_consis} presents the PCR calculation results for this benchmark.
% \usepackage{multirow}
% \usepackage{booktabs}

\begin{table}[h]
\centering
\resizebox{\columnwidth}{!}{%
\begin{tabular}{lccccc}
\toprule
\textbf{Dev Set (Dataset Size: 170)} & \multicolumn{2}{c}{Surface Level: EM} & \multicolumn{2}{c}{Fact Level: Correctness} & \multirow{2}{*}{Consistency} \\
\cmidrule(lr){2-3} \cmidrule(lr){4-5}
Model       & Original & Modified & Original & Modified &  \\
\midrule
Baichuan-13B    & 0.4235   & 0.4412   & 0.4529   & 0.4529   & 0.4765   \\
Disc-Fin-13B    & 0.4647   & 0.4059   & 0.4941   & 0.4176   & 0.4824   \\
FinMA-NLP-7B   & 0.1176   & 0.1647   & 0.2588   & 0.2941   & 0.2765   \\
FinMA-Full-7B  & 0.1824   & 0.1353   & 0.3059   & 0.2529   & 0.2706   \\
LLaMA-8B       & 0        & 0        & 0.5647   & 0.4941   & 0.6059   \\
Mistral-7B     & 0        & 0        & 0.4176   & 0.4000   & 0.5294   \\
GPT-4o-mini & 0        & 0        & 0.7059   & 0.6588   & 0.7706   \\
\midrule
\textbf{Val Set (Dataset Size: 1151)} & \multicolumn{2}{c}{Surface Level: EM} & \multicolumn{2}{c}{Fact Level: Correctness} & \multirow{2}{*}{Consistency} \\
\cmidrule(lr){2-3} \cmidrule(lr){4-5}
Model       & Original & Modified & Original & Modified &  \\
\midrule
Baichuan-13B    & 0.4544   & 0.4309   & 0.4909   & 0.4396   & 0.4674   \\
Disc-Fin-13B    & 0.4327   & 0.3884   & 0.4613   & 0.3997   & 0.4083   \\
FinMA-NLP-7B   & 0.1095   & 0.1512   & 0.3023   & 0.2789   & 0.2407   \\
FinMA-Full-7B  & 0.1706   & 0.1647   & 0.3189   & 0.2485   & 0.2502   \\
LLaMA-8B       & 0        & 0        & 0.4631   & 0.4518   & 0.5308   \\
Mistral-7B     & 0        & 0        & 0.4292   & 0.4005   & 0.4944   \\
GPT-4o-mini & 0        & 0.0009   & 0.6551   & 0.6699   & 0.7489   \\
\bottomrule
\end{tabular}%
}
\caption{FinEval Results}
\label{tab:fineval_results_all}
\end{table}

% \usepackage{multirow}
% \usepackage{graphicx}
% \usepackage{booktabs}

\begin{table}[]
\centering
\resizebox{\columnwidth}{!}{%
\begin{tabular}{lccccccc}
\toprule
\multirow{2}{*}{Model} & \multicolumn{3}{c}{Dev} & \multicolumn{3}{c}{Val} & \multirow{2}{*}{Drop} \\ 
\cmidrule(lr){2-4} \cmidrule(lr){5-7}
 & EM & Consistency & PCR & EM & Consistency & PCR &  \\ 
\midrule
Baichuan & 0.4235 & 0.4765 & 0.7119 & 0.4544 & 0.4674 & 0.7499 & \underline{-0.038} \\
Disc-Fin & 0.4647 & 0.4824 & 0.7461 & 0.4327 & 0.4083 & 0.785 & \textbf{-0.0389} \\
FinMA-NLP & 0.1176 & 0.2765 & 0.4181 & 0.1095 & 0.2407 & 0.4436 & -0.0255 \\
FinMA-Full & 0.1824 & 0.2706 & 0.5952 & 0.1706 & 0.2502 & 0.6006 & -0.0054 \\
LLaMA & 0 & 0.6059 & 0.0162 & 0 & 0.5308 & 0.0185 & -0.0023 \\
Mistral & 0 & 0.5294 & 0.0185 & 0 & 0.4944 & 0.0198 & -0.0013 \\
GPT-4o-mini & 0 & 0.7706 & 0.0128 & 0 & 0.7489 & 0.0132 & -0.0004 \\
\bottomrule
\end{tabular}%
}
\caption{PCR calculation for the FinEval benchmark using the $\tanh(\cdot)$ function and $\alpha = 0.01$}
\label{tab:fineval_consis}
\end{table}

% \usepackage{booktabs}
% \usepackage{multirow}
% \usepackage{booktabs}
% \usepackage{multirow}

\begin{table}[h]
\centering
\begin{adjustbox}{max width=\textwidth}
\begin{tabular}{@{}lccccccccc@{}}
\toprule
Model & \multicolumn{3}{c}{\textbf{Train Set (Sample Size: 500)}} & \multicolumn{3}{c}{\textbf{Validation Set (Sample Size: 500)}} & \multicolumn{3}{c}{\textbf{Test Set (Sample Size: 500)}} \\ \cmidrule(lr){2-4} \cmidrule(lr){5-7} \cmidrule(lr){8-10}
& Original & Modified & Consistency & Original & Modified & Consistency & Original & Modified & Consistency \\ \midrule
LLaMA-8B & 0.0021 & 0.0022 & 0.8326 & 0.0021 & 0.0022 & 0.83 & 0.0022 & 0.0024 & 0.8246 \\
Mistral-7B & 0.0209 & 0.022 & 0.5554 & 0.0229 & 0.0225 & 0.5679 & 0.0201 & 0.0205 & 0.5656 \\
FinMA-Full-7B & 0.2304 & 0.2181 & 0.727 & 0.0943 & 0.0835 & 0.5662 & 0.1084 & 0.1016 & 0.5706 \\
FinMA-NLP-7B & 0.1481 & 0.1391 & 0.6993 & 0.092 & 0.0866 & 0.638 & 0.106 & 0.106 & 0.6219 \\
GPT-4o-mini & 0.0209 & 0.0217 & 0.4562 & 0.0217 & 0.0206 & 0.4604 & 0.0208 & 0.0207 & 0.4505 \\ \midrule
\multicolumn{1}{l}{} & \multicolumn{3}{c}{Train Set (Sample Size: 100)} & \multicolumn{3}{c}{Validation Set (Sample Size: 100)} & \multicolumn{3}{c}{Test Set (Sample Size: 100)} \\ \cmidrule(lr){2-4} \cmidrule(lr){5-7} \cmidrule(lr){8-10}
Baichuan-13B & 0.0084 & 0.0078 & 0.2506 & 0.01 & 0.0131 & 0.2546 & 0.0081 & 0.0068 & 0.2466 \\
Disc-Fin-13B & 0.0097 & 0.009 & 0.512 & 0.0118 & 0.0135 & 0.5554 & 0.0144 & 0.0159 & 0.4796 \\ 
\bottomrule
\end{tabular}
\end{adjustbox}
\caption{FinQA Results}
\label{tab:finqa_results_all}
\end{table}

% \usepackage{booktabs}
% \usepackage{multirow}
% \usepackage{graphicx}
% \usepackage[normalem]{ulem}

\begin{table*}[ht]
\centering
\resizebox{\textwidth}{!}{%
\begin{tabular}{lrrrrrrrrrrr}
\toprule
\multirow{2}{*}{Model} & \multicolumn{3}{c}{Train} & \multicolumn{3}{c}{Validation} & \multicolumn{3}{c}{Test} & \multirow{2}{*}{Drop\_Train\_Val} & \multirow{2}{*}{Drop\_Train\_Test} \\ 
\cmidrule(lr){2-4} \cmidrule(lr){5-7} \cmidrule(lr){8-10}
 & Rouge-L & Consistency & PCR & Rouge-L & Consistency & PCR & Rouge-L & Consistency & PCR &  &  \\
\midrule
LLaMA-8B & 0.0021 & 0.8326 & 0.0144 & 0.0021 & 0.83 & 0.0144 & 0.0022 & 0.8246 & 0.0147 & -0.0001 & -0.0003 \\
Mistral-7B & 0.0209 & 0.5554 & 0.0546 & 0.0229 & 0.5679 & 0.0569 & 0.0201 & 0.5656 & 0.0523 & -0.0023 & 0.0023 \\
FinMA-Full-7B & 0.2304 & 0.727 & 0.3151 & 0.0943 & 0.5662 & 0.1791 & 0.1084 & 0.5706 & 0.2011 & \textbf{0.136} & \textbf{0.1139} \\
FinMA-NLP-7B & 0.1481 & 0.6993 & 0.2193 & 0.092 & 0.638 & 0.1561 & 0.1060 & 0.6219 & 0.1816 & {\underline{0.0633}} & \underline{0.0377} \\
GPT-4o-mini & 0.0209 & 0.4562 & 0.0662 & 0.0217 & 0.4604 & 0.0672 & 0.0208 & 0.4505 & 0.0667 & -0.001 & -0.0005 \\
Baichuan-13B & 0.0084 & 0.2506 & 0.0703 & 0.01 & 0.2546 & 0.0756 & 0.0081 & 0.2466 & 0.0704 & -0.0053 & -0.0001 \\
Disc-Fin-13B & 0.0097 & 0.512 & 0.0377 & 0.0118 & 0.5554 & 0.0385 & 0.0144 & 0.4796 & 0.0497 & -0.0009 & -0.0121 \\
\bottomrule
\end{tabular}
}
\caption{PCR calculation for the FinQA benchmark using the $\tanh(\cdot)$ function and $\alpha = 0.01$}
\label{tab:finqa_consis}
\end{table*}

% \usepackage{booktabs}
\begin{table}[h]
\centering
\begin{tabular}{@{}lcccccccc@{}}
\toprule
              & \multicolumn{8}{c}{\textbf{Train Set (Sample Size: 500)}}                                            \\ \cmidrule(lr){2-9}
              & \multicolumn{2}{c}{BLEU} & \multicolumn{2}{c}{Rouge-1} & \multicolumn{2}{c}{Rouge-2} & \multicolumn{2}{c}{Rouge-L} \\ 
\cmidrule(lr){2-3} \cmidrule(lr){4-5} \cmidrule(lr){6-7} \cmidrule(lr){8-9}
              & Original    & Modified   & Original     & Modified     & Original     & Modified     & Original     & Modified     \\ 
\midrule
LLaMA-8B      & 0.0002      & 0.0002     & 0.0089       & 0.009        & 0.0009       & 0.0009       & 0.0088       & 0.0089       \\
Mistral-7B    & 0.0047      & 0.0047     & 0.126        & 0.1275       & 0.0205       & 0.0204       & 0.1252       & 0.1267       \\
FinMA-Full-7B & 0.048       & 0.048      & 0.779        & 0.745        & 0.242        & 0.232        & 0.779        & 0.745        \\
FinMA-NLP-7B  & 0.048       & 0.048      & 0.5221       & 0.5188       & 0.192        & 0.188        & 0.5221       & 0.5188       \\
Baichuan-13B  & 0.0038      & 0.0038     & 0.1128       & 0.1106       & 0.0093       & 0.0092       & 0.1127       & 0.1105       \\
Disc-Fin-13B  & 0.0043      & 0.0043     & 0.0883       & 0.0877       & 0.0073       & 0.0075       & 0.0883       & 0.0877       \\
GPT-4o-mini   & 0.0221      & 0.0195     & 0.2548       & 0.2599       & 0.0406       & 0.0366       & 0.2545       & 0.2591       \\
\midrule
              & \multicolumn{8}{c}{\textbf{Validation Set (Sample Size: 500)}}                                       \\ \cmidrule(lr){2-9}
              & \multicolumn{2}{c}{BLEU} & \multicolumn{2}{c}{Rouge-1} & \multicolumn{2}{c}{Rouge-2} & \multicolumn{2}{c}{Rouge-L} \\
\cmidrule(lr){2-3} \cmidrule(lr){4-5} \cmidrule(lr){6-7} \cmidrule(lr){8-9}
              & Original    & Modified   & Original     & Modified     & Original     & Modified     & Original     & Modified     \\
\midrule
LLaMA-8B      & 0.0003      & 0.0003     & 0.0095       & 0.009        & 0.001        & 0.001        & 0.0093       & 0.0088       \\
Mistral-7B    & 0.0024      & 0.0024     & 0.1263       & 0.1293       & 0.0122       & 0.0126       & 0.1245       & 0.1275       \\
FinMA-Full-7B & 0.0474      & 0.0474     & 0.5795       & 0.5842       & 0.1514       & 0.1494       & 0.5792       & 0.5839       \\
FinMA-NLP-7B  & 0.0566      & 0.0566     & 0.4399       & 0.4385       & 0.1482       & 0.1482       & 0.4395       & 0.4381       \\
Baichuan-13B  & 0.003       & 0.003      & 0.0953       & 0.0904       & 0.0089       & 0.0091       & 0.0951       & 0.0902       \\
Disc-Fin-13B  & 0.0012      & 0.0012     & 0.0798       & 0.0789       & 0.0121       & 0.0121       & 0.0798       & 0.0789       \\
GPT-4o-mini   & 0.01        & 0.0109     & 0.2563       & 0.2702       & 0.0195       & 0.0207       & 0.2556       & 0.27         \\
\bottomrule
\end{tabular}
\caption{AlphaFin Results}
\label{tab:alphafin_results_all}
\end{table}

\begin{table}[]
\centering
\resizebox{\columnwidth}{!}{%
\begin{tabular}{lccccccc}
\toprule
\multirow{2}{*}{Model} & \multicolumn{3}{c}{Train} & \multicolumn{3}{c}{Validation} & \multirow{2}{*}{Drop} \\ 
\cmidrule(lr){2-4} \cmidrule(lr){5-7}
 & Rouge-L & Consistency & PCR & Rouge-L & Consistency & PCR &  \\ 
\midrule
LLaMA & 0.0088 & 0.7919 & 0.0234 & 0.0093 & 0.7702 & 0.0247 & -0.0013 \\
Mistral & 0.1252 & 0.6838 & 0.1924 & 0.1245 & 0.6581 & 0.1987 & -0.0062 \\
FinMA-Full & 0.7790 & 0.8964 & 0.7016 & 0.5792 & 0.8884 & 0.5756 & \textbf{0.1261} \\
FinMA-NLP & 0.5221 & 0.8873 & 0.5321 & 0.4395 & 0.8905 & 0.4615 & \underline{0.0706} \\
Baichuan & 0.1127 & 0.5760 & 0.2063 & 0.0951 & 0.5513 & 0.1851 & 0.0213 \\
Disc-Fin & 0.0883 & 0.7381 & 0.1307 & 0.0798 & 0.7321 & 0.1204 & 0.0103 \\
GPT-4o-mini & 0.2545 & 0.5417 & 0.4457 & 0.2556 & 0.5472 & 0.4436 & 0.0022 \\
\bottomrule
\end{tabular}%
}
\caption{PCR calculation for the AlphaFin benchmark using the $\tanh(\cdot)$ function and $\alpha = 0.01$}
\label{tab:alphafin_consis}
\end{table}

\begin{table}[h]
\centering
\begin{tabular}{llcccccc}
\toprule
\multirow{2}{*}{Model} & \multirow{2}{*}{\begin{tabular}[c]{@{}c@{}}Sample\\ Size\end{tabular}} &
\multicolumn{2}{c}{ROUGE-1} & \multicolumn{2}{c}{ROUGE-2} & \multicolumn{2}{c}{ROUGE-L} \\
\cmidrule(r){3-4} \cmidrule(r){5-6} \cmidrule(r){7-8}
                      &                      & Original & Modified & Original & Modified & Original & Modified \\
\midrule
\textbf{Train}        &                      &          &          &          &          &          &          \\
LLaMA-8B                 & 100                  & 0.0709   & 0.0713   & 0.0281   & 0.0257   & 0.0505   & 0.0482   \\
Mistral-7B               & 517                  & 0.131    & 0.1247   & 0.0493   & 0.0441   & 0.0834   & 0.0775   \\
FinMA-NLP-7B            & 517                  & 0.0288   & 0.031    & 0.0013   & 0.0015   & 0.0269   & 0.0279   \\
FinMA-Full-7B           & 517                  & 0.0277   & 0.0361   & 0.0024   & 0.0023   & 0.0248   & 0.0317   \\
Baichuan-13B              & 100                  & 0.0485   & 0.0441   & 0.0067   & 0.0043   & 0.0359   & 0.0329   \\
Disc-Fin-13B             & 517                  & 0.026    & 0.0283   & 0.0038   & 0.0053   & 0.0241   & 0.0261   \\
GPT-4o-mini                   & 517                  & 0.1472   & 0.1437   & 0.0467   & 0.0435   & 0.0898   & 0.0868   \\
\midrule
\textbf{Validation}   &                      &          &          &          &          &          &          \\
LLaMA-8B                 & 100                  & 0.0676   & 0.0642   & 0.0253   & 0.0229   & 0.0475   & 0.0448   \\
Mistral-7B               & 101                  & 0.1391   & 0.129    & 0.048    & 0.0448   & 0.0855   & 0.0768   \\
FinMA-NLP-7B            & 101                  & 0.0185   & 0.0275   & 0.0001   & 0.0008   & 0.0173   & 0.0252   \\
FinMA-Full-7B           & 101                  & 0.0311   & 0.0318   & 0.002    & 0.0025   & 0.0279   & 0.0275   \\
Baichuan-13B              & 100                  & 0.0559   & 0.0492   & 0.0074   & 0.0057   & 0.0387   & 0.0345   \\
Disc-Fin-13B             & 101                  & 0.0237   & 0.0176   & 0.0034   & 0.0038   & 0.022    & 0.0168   \\
GPT-4o-mini                   & 101                  & 0.1499   & 0.1511   & 0.0456   & 0.0467   & 0.0884   & 0.0887   \\
\midrule
\textbf{Test}         &                      &          &          &          &          &          &          \\
LLaMA-8B                 & 100                  & 0.06     & 0.0594   & 0.0221   & 0.021    & 0.0411   & 0.0407   \\
Mistral-7B               & 250                  & 0.1222   & 0.1205   & 0.0443   & 0.0411   & 0.0784   & 0.0754   \\
FinMA-NLP-7B            & 250                  & 0.0284   & 0.0288   & 0.0011   & 0.0013   & 0.0261   & 0.0265   \\
FinMA-Full-7B           & 250                  & 0.031    & 0.0343   & 0.0023   & 0.0021   & 0.0264   & 0.03     \\
Baichuan-13B              & 100                  & 0.0413   & 0.0462   & 0.0044   & 0.0054   & 0.0315   & 0.0329   \\
Disc-Fin-13B             & 250                  & 0.0308   & 0.0238   & 0.0055   & 0.0047   & 0.0284   & 0.0226   \\
GPT-4o-mini                   & 250                  & 0.1413   & 0.1373   & 0.0419   & 0.0409   & 0.0854   & 0.0826   \\
\bottomrule
\end{tabular}
\caption{ECTSum Results}
\label{tab:ectsum_results_all}
\end{table}

\begin{table*}[t]
\centering
\resizebox{\textwidth}{!}{%
\begin{tabular}{lrrrrrrrrrrr}
\toprule
\multirow{2}{*}{Model} & \multicolumn{3}{c}{Train} & \multicolumn{3}{c}{Validation} & \multicolumn{3}{c}{Test} & \multirow{2}{*}{Drop\_Train\_Val} & \multirow{2}{*}{Drop\_Train\_Test} \\ 
\cmidrule(lr){2-4} \cmidrule(lr){5-7} \cmidrule(lr){8-10}
 & Rouge-2 & Consistency & PCR & Rouge-2 & Consistency & PCR & Rouge-2 & Consistency & PCR &  &  \\
\midrule
LLaMA-8B      & 0.0281   & 0      & 1  & 0.0253   & 0.0200      & 0.8491  & 0.0221    & 0      & 1  & \textbf{0.1509}   & 0   \\
Mistral-7B    & 0.0493   & 0.0116      & 0.9993  & 0.0480   & 0.0099      & 0.9998  & 0.0443    & 0.016      & 0.9904  & -0.0004  & 0.0089   \\
FinMA-Full-7B & 0.0024   & 0.029      & 0.1119  & 0.002   & 0.0396      & 0.0738  & 0.0023    & 0.036      & 0.0903  & \underline{0.0382}   & \underline{0.0216}   \\
FinMA-NLP-7B  & 0.0013   & 0.0445      & 0.0507  & 0.0001   & 0.0396      & 0.0272  & 0.0011    & 0.028      & 0.0730  & 0.0235   & \textbf{-0.0223}  \\
Baichuan-13B  & 0.0067   & 0      & 1  & 0.0074   & 0      & 1  & 0.0044    & 0      & 1  & 0   & 0   \\
Disc-Fin-13B  & 0.0038   & 0.0522      & 0.0907  & 0.0034   & 0.0693      & 0.0632  & 0.0055    & 0.0600      & 0.1054  & 0.0276   & -0.0147  \\
GPT-4o-mini   & 0.0467   & 0.0077      & 1  & 0.0456   & 0.0099      & 0.9996  & 0.0419    & 0.012      & 0.9973  & 0.0004   & 0.0027   \\
\bottomrule
\end{tabular}%
}
\caption{PCR calculation for the ECTSum benchmark using the $\tanh(\cdot)$ function and $\alpha = 0.001$}
\label{tab:ectsum_consis}
\end{table*}

\section{Sample Appendix}
\label{sec:appendix_sample}
\begin{figure*}[h]
    \centering
    \begin{subfigure}[a]{\textwidth}
        \includegraphics[width=\textwidth]{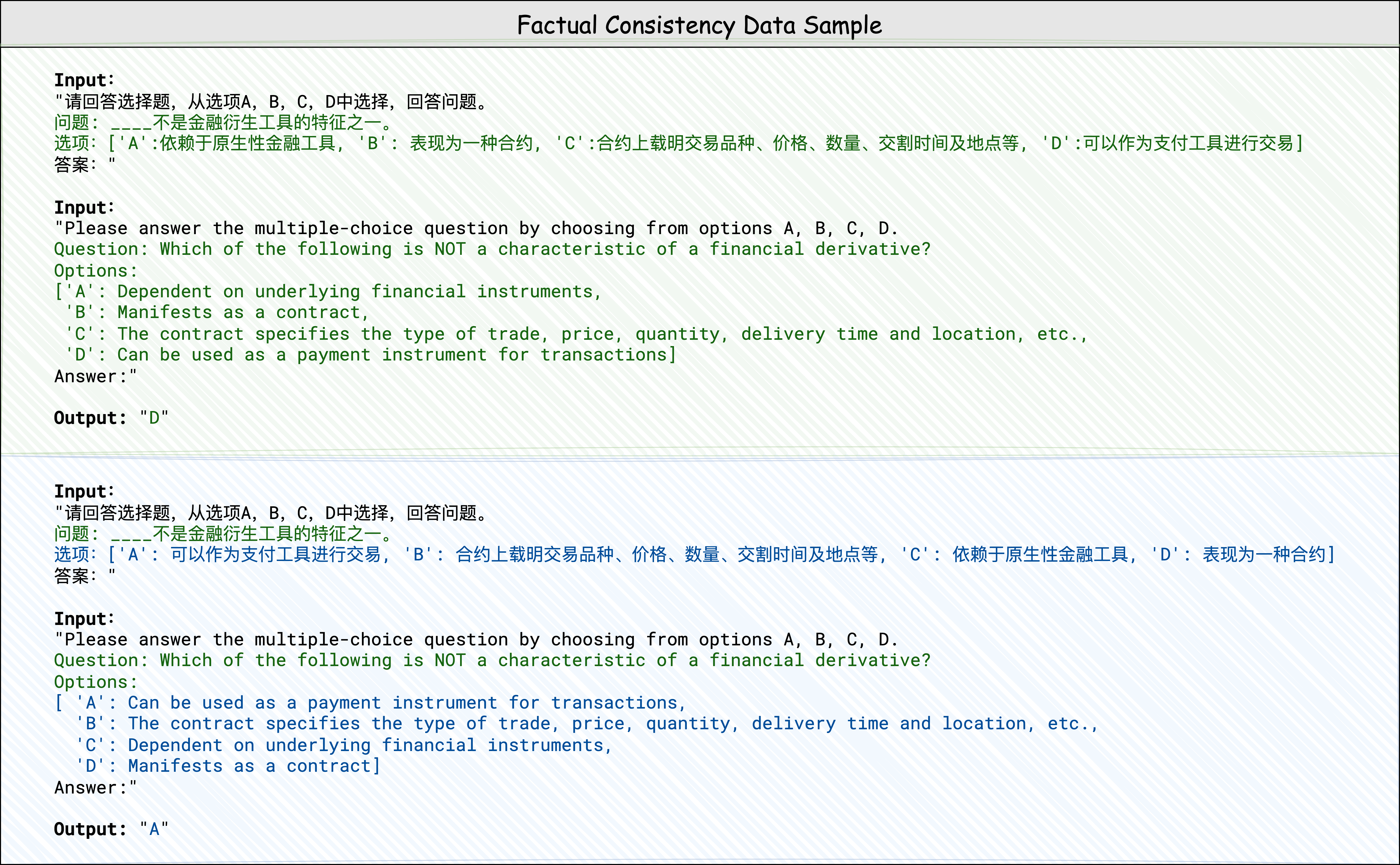}
        \caption{Sample of Factual-Consistency-Based Data Modification}
        \label{fig:sample_factual}
    \end{subfigure}

    \begin{subfigure}[b]{\textwidth}
        \includegraphics[width=\textwidth]{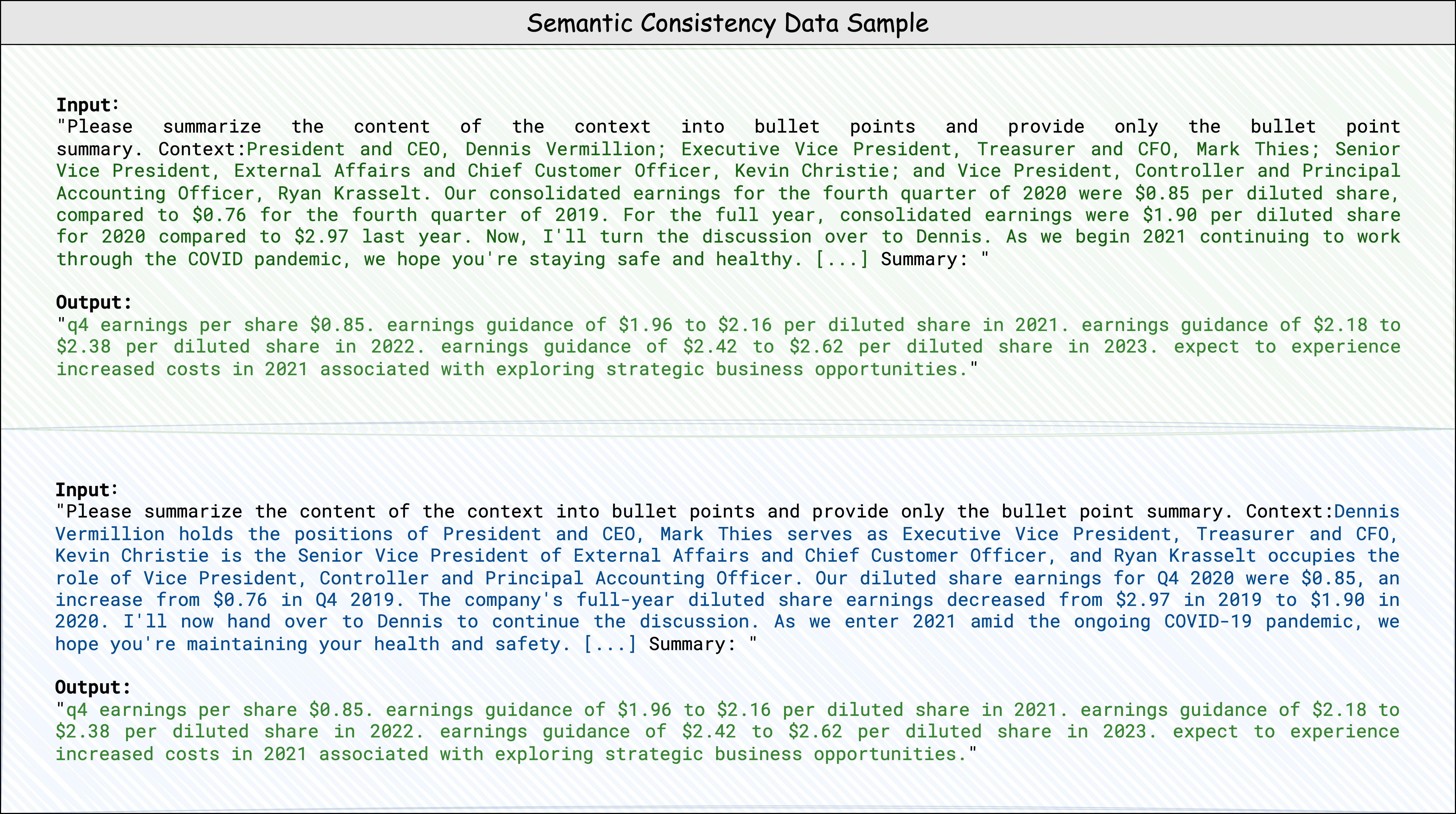}
        \caption{Sample of Semantic-Consistency-Based Data Modification}
        \label{fig:sample_semantic}
    \end{subfigure}
    \caption{Samples of Factual-Consistency-Based and Semantic-Consistency-Based Data Modification}
    \label{fig:combined_samples}
\end{figure*}

\begin{figure*}[h]
    \centering
    \begin{subfigure}[a]{\textwidth}
        \includegraphics[width=\textwidth]{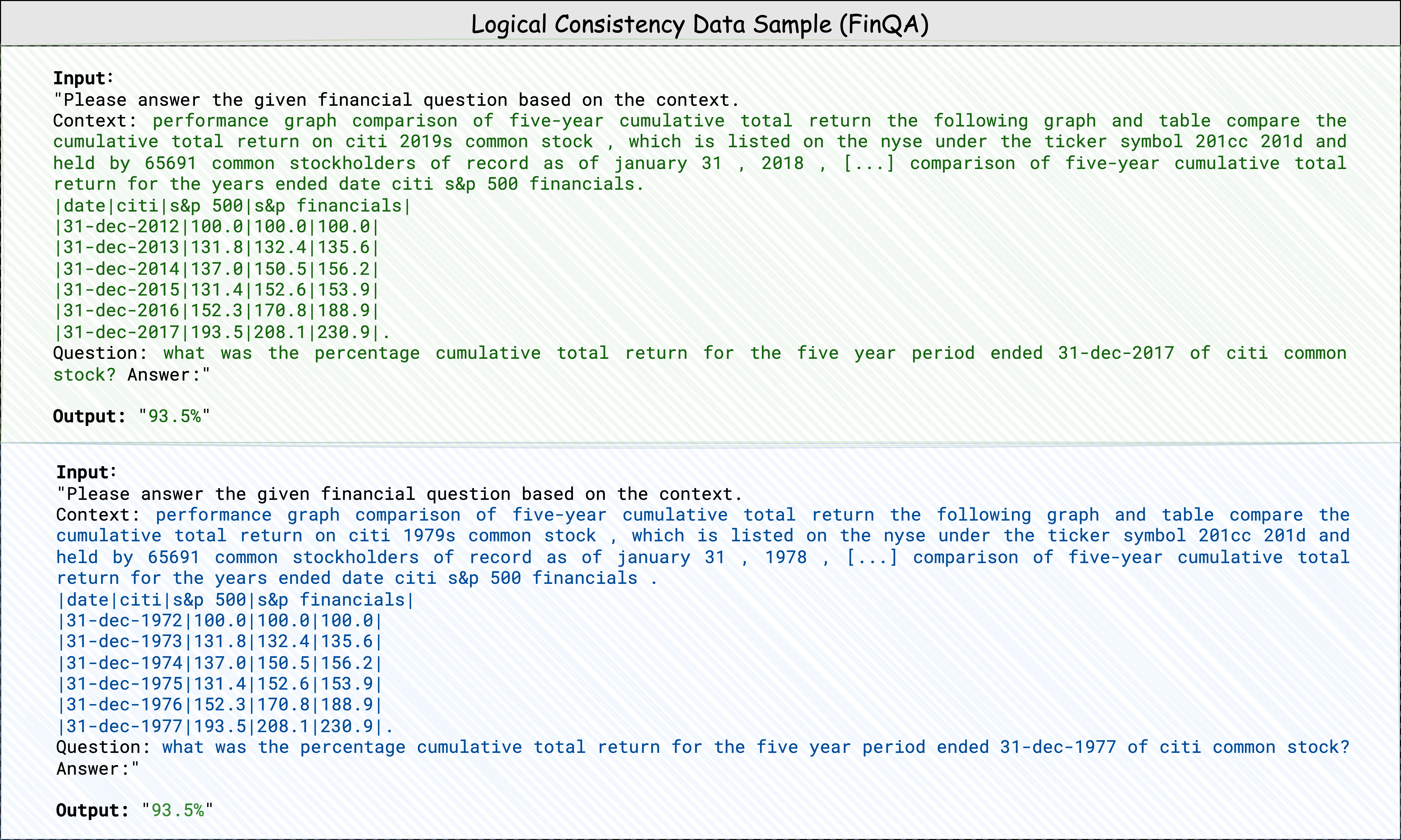}
        \caption{Sample of Logical-Consistency-Based Data Modification (FinQA)}
        \label{fig:sample_log_1}
    \end{subfigure}

    \begin{subfigure}[b]{\textwidth}
        \includegraphics[width=\textwidth]{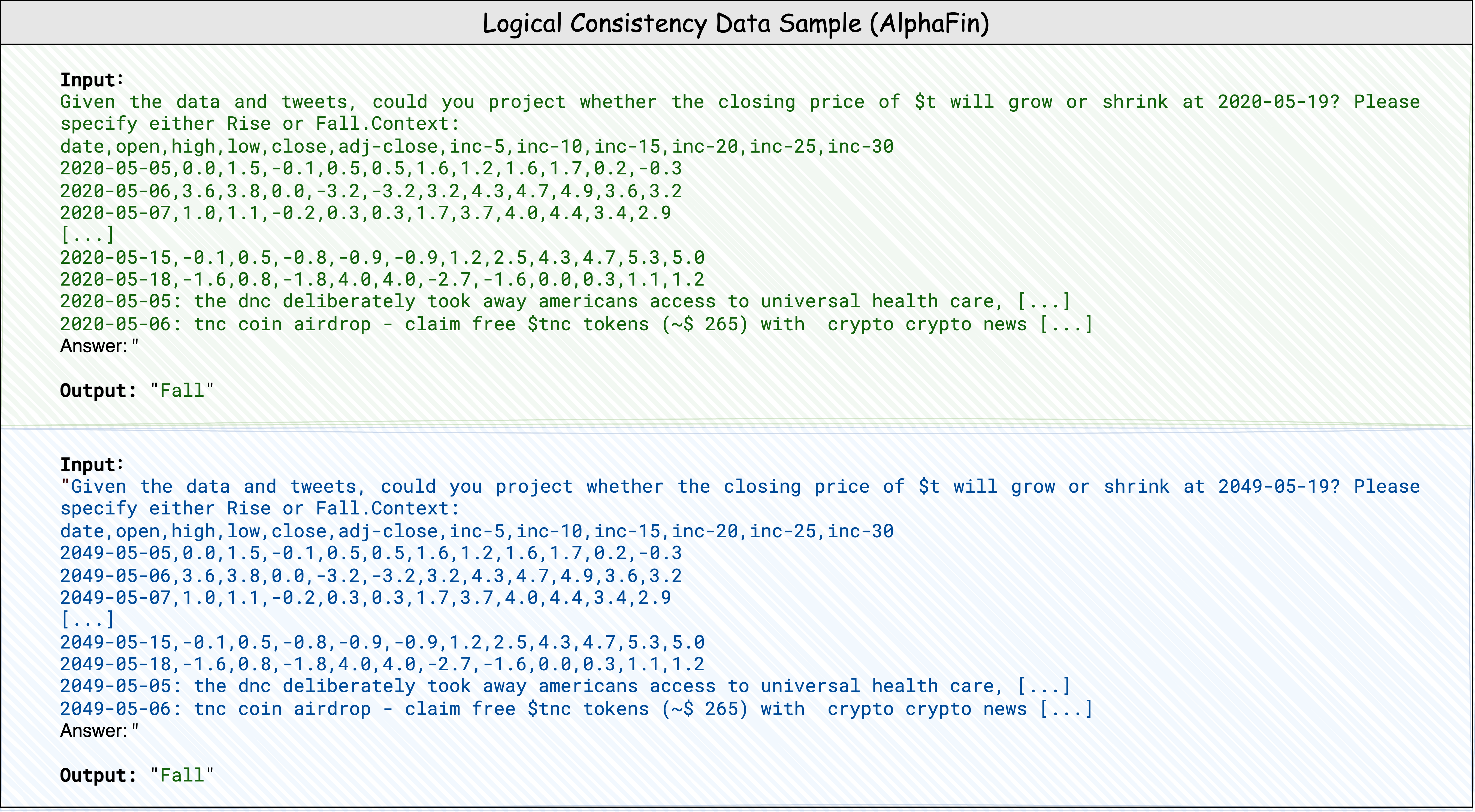}
        \caption{Sample of Logical-Consistency-Based Data Modification (AlphaFin)}
        \label{fig:sample_log_2}
    \end{subfigure}
    \caption{Samples of Logical-Consistency-Based Data Modification}
    \label{fig:combined_samples_2}
\end{figure*}

Figures \ref{fig:combined_samples} and \ref{fig:combined_samples_2} present the original and modified data samples for each benchmark.

\section{Prompt Appendix}
\label{sec:appendix_prompt}
Table \ref{tab:prompt_ectsum} shows the prompt used with \texttt{claude-3.5-sonnet-20240620} to generate semantically equivalent text for the ECTSum dataset.
\begin{table*}
\centering
    \begin{tabularx}{\textwidth}{l|X}
        \toprule
        \textbf{Prompt} & Your task is to paraphrase sentences. The input consists of sentences separated by `\textbackslash n.' You need to rewrite each sentence in different words while retaining the same meaning. Then, you should output only the paraphrased sentences. Remember not to include phrases like "the rewritten sentences are..." Also, do not include indexes like 1, 2, 3; just separate them with `\textbackslash n.' \\ 
        & \\
        & The input sentences are: \\
        & \textit{Second quarter adjusted income from continuing operations per diluted share increased to \$0.82, up nearly 37\% from the year ago quarter.}\\
        & \textit{We generated significant operating leverage with adjusted EBITDA improving 17\% year-over-year to \$106.6 million and adjusted EBITDA margin increasing 100 basis points to 7.1\% on slightly higher revenues.} \\
        & ... \\ 
        & \\
        & The output is: \\
        
        \midrule
        \textbf{Output} & \textit{Adjusted income from ongoing operations per diluted share in the second quarter rose to \$0.82, an increase of almost 37\% compared to the same period last year.}
        
        \textit{We achieved considerable operational efficiency, with adjusted EBITDA growing 17\% year-over-year to \$106.6 million and adjusted EBITDA margin expanding 100 basis points to 7.1\% on marginally higher revenues.} \\
        & ...\\
        \bottomrule
    \end{tabularx}
    \caption{Prompt for generating paraphrased texts with the Claude model, version \texttt{claude-3.5-sonnet-20240620}.}
    \label{tab:prompt_ectsum}
\end{table*}

\begin{table}[]
\centering
\resizebox{0.4\columnwidth}{!}{%
\begin{tabular}{llll}
\toprule
 & Train & Val & Test \\
 \midrule
Min & 0.9791 & 0.9804 & 0.9844 \\
Mean & 0.9984 & 0.9984 & 0.9979 \\
Max & 0.9999 & 0.9999 & 0.9999 \\
\bottomrule
\end{tabular}%
}
\caption{Cosine Similarity between Original and Modified ECTSum Texts using Longformer Encoding.}
\label{tab:ectsum_sim}
\end{table}
\begin{figure}[ht]
    \centering
        \centering
        \includegraphics[width=\columnwidth]{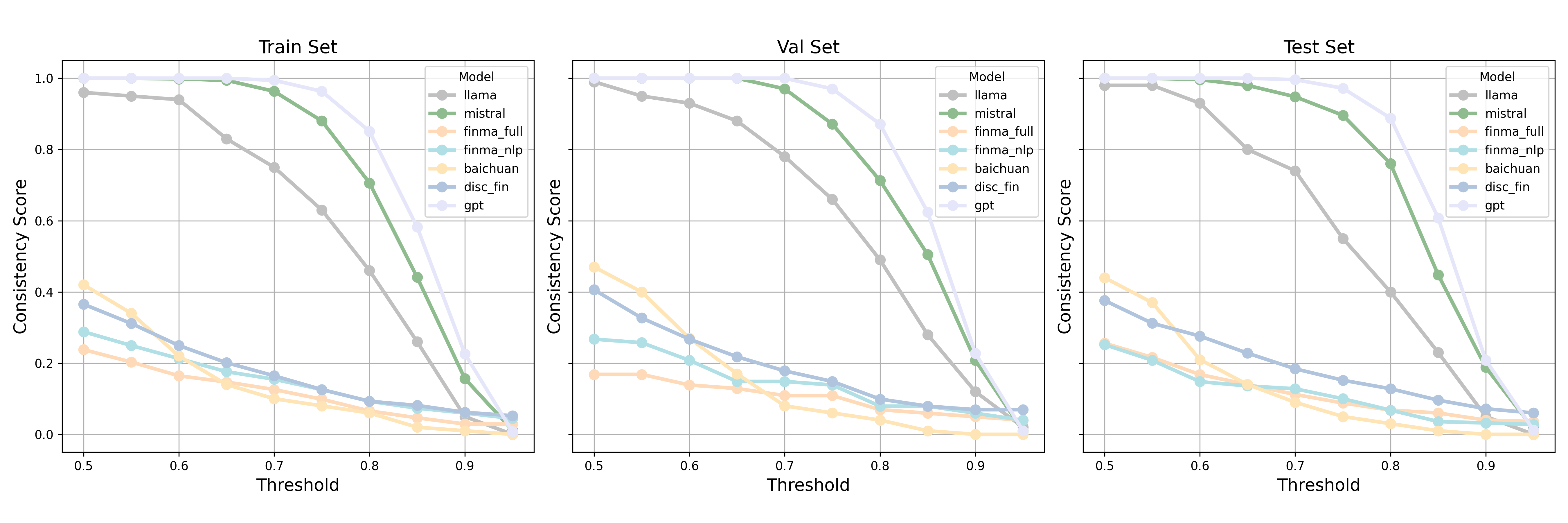}
        \caption{The consistency of LLMs' generation before and after input text modification varies with changes in the similarity threshold. As the threshold increases, the consistency decreases.}
        \label{fig:ectsum_sim_thre}
\end{figure}
\section{Overlap Appendix}
\label{sec:appendix_overlap}
\begin{figure*}[h]
    \centering
        \includegraphics[width=\textwidth]{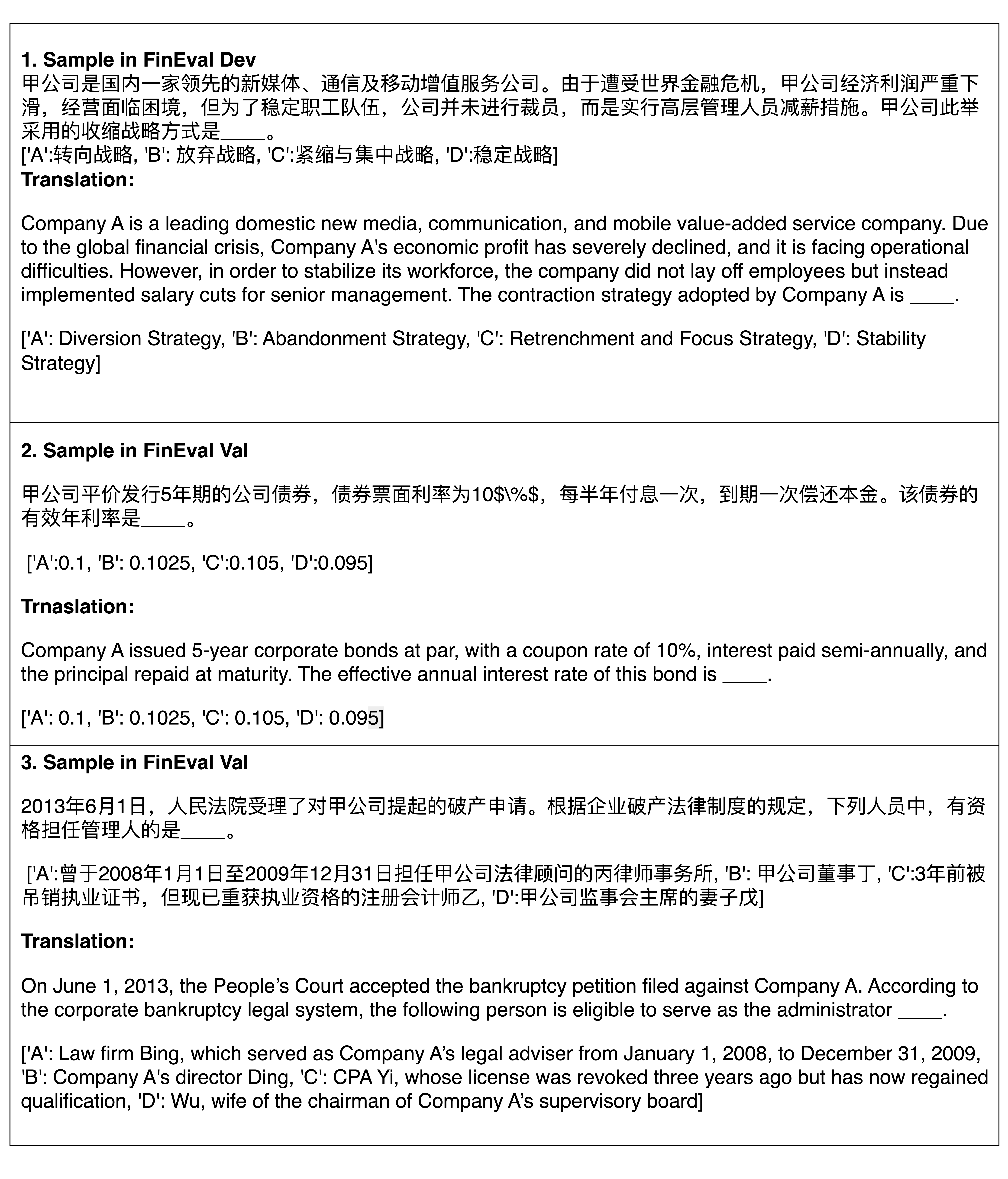}
        \caption{Benchmark Overlap}
        \label{fig:benchmark_overlap}
\end{figure*}

Examples of benchmark overlap between FinEval and C-Eval are shown in Figure~\ref{fig:benchmark_overlap}.

\end{document}